\documentclass[11pt]{article}
\usepackage{epsfig}
\usepackage{amssymb,amsmath}
\usepackage{epigraph}

\usepackage{graphicx}				
\usepackage{amssymb}
\usepackage{bm}

\def\xdot{\overset{\bm .}{{\bf x}}}

\def\boldmu{\text{\boldmath$\mu$}}
\def\boldsigma{\text{\boldmath$\sigma$}}

\newcommand{\beq}{\begin{equation}}
\newcommand{\eeq}{\end{equation}}
\newcommand{\bea}{\begin{eqnarray}}
\newcommand{\eea}{\end{eqnarray}}

\begin{document}



\begin{center}

{\LARGE
 Distributional Offline Continuous-Time Reinforcement Learning
\vskip0.5cm
with Neural Physics-Informed PDEs (SciPhy RL for DOCTR-L)
}

\vskip1.0cm
{\Large Igor Halperin} \\
\vskip0.5cm
AI AM Research, Fidelity Investments\\
\vskip0.5cm
{\small e-mail: $ighalp@gmail.com $}
\vskip0.5cm
\today \\

\vskip1.0cm
{\Large Abstract:\\}
\end{center}
\parbox[t]{\textwidth}{
This paper addresses  distributional offline continuous-time reinforcement learning (DOCTR-L) with stochastic policies for high-dimensional optimal control.
A soft distributional version of the classical Hamilton-Jacobi-Bellman (HJB) equation is given by 
a semilinear partial differential equation (PDE). This `soft HJB equation' can be \emph{learned} from offline data \emph{without} assuming that the latter 
correspond to a previous optimal or near-optimal policy.
A data-driven solution of the soft HJB equation uses methods of Neural PDEs and Physics-Informed Neural Networks developed in the field of Scientific Machine Learning (SciML). The suggested approach, dubbed `SciPhy RL', thus reduces DOCTR-L to solving neural PDEs from data. Our algorithm called Deep DOCTR-L converts offline high-dimensional data into an optimal policy in \emph{one step} by reducing it to supervised learning, instead of relying on value iteration or policy iteration methods. The method enables a computable approach to the quality control of obtained policies in terms of both expected returns and uncertainties about their values. 
}

    \newcounter{helpfootnote}
\setcounter{helpfootnote}{\thefootnote} 
\renewcommand{\thefootnote}{\fnsymbol{footnote}}
\setcounter{footnote}{0}
\footnotetext{
}     

 \renewcommand{\thefootnote}{\arabic{footnote}}
\setcounter{footnote}{\thehelpfootnote} 

\newpage
 
\section{Introduction}

Reinforcement learning (RL) provides a framework for data-driven, learning-based approaches to problems of optimal control \cite{Sutton_Barto_2018}. 
In addition in relying on data and relaxing the dependence on a model for dynamics of an environment, RL also offers new computational methods - which becomes especially 
important for many real-life problems of high-dimensional optimal control. For such settings, 
classical methods based on the Bellman equation for discrete-time problems or the Hamilton-Jacobi-Bellman (HJB) equation for continuous-time problems (see e.g. \cite{Bertsekas_2019}) become computationally infeasible.
A tremendous success was achieved in the recent years with using RL methods for many high-dimensional optimal control problems, 
including e.g. achieving a super-human performance in the game of Go \cite{Silver_2017}.
These approaches are generally known as deep RL, and are based on a combination of methods of RL with deep neural networks to provide flexible function approximations for learning. 

Most of existing RL or deep RL algorithms are \emph{online} methods where an agent has access to its environment, and can explore different policies. This paper addresses \emph{offline} RL
(also known as batch-mode RL), 
where the agent only has the ability to utilize previously collected \emph{offline data}, but cannot be engaged in any additional online interaction with the environment for the purpose of training. The offline data may be collected from actions of another agent, whose objective could be altogether different from the objective of the first agent. 

Clearly, assumptions of offline RL fit very well many real-world problems where previously collected historical data could be utilized to optimize decision-making going forward. Offline RL attracted a lot of interest in the research community, see \cite{Levine_Offline_RL_2020} for a review and references to the original literature.  As was emphasized in  \cite{Levine_Offline_RL_2020}, while supervised learning provides a framework for data-based \emph{pattern recognition}, offline RL provides a framework for data-driven \emph{decision-making}, and therefore is very appealing and universal.
On the other hand, by the very nature of its task, offline RL is also \emph{harder} than the traditional online RL. This is because the agent cannot explore states and actions that are not encountered in the training data. This may produce \emph{exploration errors} \cite{Fujimoto_2019}, where a RL model, unless properly tamed, can make erroneous and unrealistic estimates for unseen state-action pairs.
 
This paper considers a continuous-time formulation of offline RL. Continuous-time RL (CTRL) with continuous state and action spaces and \emph{deterministic} policies provides a data-driven framework for solving the HJB equation \cite{Doya_2000}. Partial differential equations (PDEs) such as the HJB equation encode information about derivatives of a cost function, and hence potentially on values of state-action pairs not seen in the training data. The notion of continuity is therefore naturally embedded in CTRL, which makes it a particularly attractive formulation for offline RL as a principled way to tame exploration errors.   

Instead of working with deterministic policies as in  \cite{Doya_2000}, here we address CTRL with \emph{stochastic} policies by adapting ideas from (discrete-time) Maximum Entropy RL (MaxEnt RL) (see e.g. \cite{Levine_2018} for a review) to a continuous-time setting. On top of that, the method presented in this paper aims at control of the whole \emph{distribution} of future returns, and not only the \emph{expected} future returns. Note that traditional RL methods only optimize the expected returns, but do not control risk (uncertainty) of these returns. For this reason, they are sometimes referred as `risk-neutral' RL methods. Unlike the latter, distributional RL \cite{Bellemare_2017} and risk-sensitive RL \cite{Shen_2014} aim to also control \emph{uncertainties} (risk) of returns. These approaches are typically based on distributional versions of time difference (TD) methods. 

This paper provides a different, probabilistic approach to control of uncertainty of returns in continuous time with offline RL (Distributional Offline Continuous Time RL, or DOCTR-L for short) that is based on PDEs. We note here a recent work in \cite{Wang_2020} that developed a continuous time MaxEnt RL formulation for the conventional `risk-neutral' RL setting.
Another difference between this work and  \cite{Wang_2020} is that while the authors in  \cite{Wang_2020}  focused on analytically solvable cases such as a continuous-time entropy-regularized linear quadratic regulator (LQR), here we are interested in a more general  high-dimensional setting, where one should rely on numerical methods. 

Expanding on this previous research, here we pursue DOCTR-L for  high-dimensional stochastic non-linear systems.  
As will be shown below, using flexible parameterizations of stochastic policies via Gaussian mixtures, the problem of DOCTR-L can be reduced to
a soft (and distributional) generalization of the HJB equation, which we call the `soft HJB equation'.  We develop a method to \emph{learn} the soft HJB equation and the optimal policy from 
offline data \emph{without} assuming that the data correspond to an optimal or near-optimal behavior of an agent that collected the data. 
A data-driven solution of the soft HJB equation relies on methods of Neural PDEs  and Physics-Informed Neural Networks (PINNs), developed in the burgeoning field of Scientific Machine Learning (SciML). The suggested approach, dubbed 'SciPhy RL', thus \emph{reduces distributional offline continuous-time RL to 
learning neural PDEs from data}. Our algorithm called Deep DOCTR-L works in high dimensions, and enables a computable approach to the quality control of obtained policies in terms of both their expected returns and uncertainties about these values. 

The rest of the paper is organized as follows. Subsection~\ref{sect_Contribution} summarizes contributions of this work.
The next subsection~\ref{sect_Related_work} provides a brief overview of related previous research.
 Sect.~\ref{Soft_HJB_control} derives the semilinear PDE (the 'soft HJB' equation) for 
the distributional continuous-time control problem. 
Sect.~\ref{sect_Neural_PDEs} shows how the soft HJB equation can be learned from behavioral data using neural PDEs with the Deep DOCTR-L algorithm. Sect.~\ref{sect_Experiments} considers numerical experiments in a 10-dimensional and 100-dimensional state spaces. Finally, Sect.~\ref{sect_Summary} provides a summary.

\subsection{Contributions of this work}
\label{sect_Contribution}

This work makes the following contributions:
\begin{itemize}
\item It proposes a new approach for \emph{control of uncertainty} of total rewards for a finite-horizon Distributional Offline Continuous Time Reinforcement Learning (DOCTR-L). Unlike other distributional RL or risk-sensitive RL approaches which are usually based on sample-based TD methods (and are typically employed for infinite-horizon problems), here we suggest a \emph{probabilistic} approach that relies on backward PDEs. This enables using numerical methods developed for PDEs for problems that are amenable to methods of DOCTR-L. 
\item With the suggested approach, control of the whole return \emph{distribution}  is computationally no harder than control of only \emph{expected} returns as done in the traditional 'risk-neutral' RL. This is unlike most of other methods of risk-sensitive RL or distributional RL, which are usually more computationally demanding than risk-neutral RL approaches.
 \item We derive a soft relaxation of the classical HJB equation (a '\emph{soft HJB equation}') for continuous-time RL with stochastic policies in the setting of offline distributional learning.
 \item The continuous-time approach adopted in this paper offers a way of containing exploration errors in offline RL using \emph{regularization by PDEs}, which enforces smoothness of resulting value functions across a fixed dataset available for training and 'imaginary' data envisioned by the model within a policy
 optimization algorithm.      
\item As a by-product, we also derive a backward (Kolmogorov) equation for the conditional probability of total reward, that depends on the agent's policy.
Solving this equation for different suggested policies enables a detailed quantitative analysis of their impact on both expected costs of different policies, and uncertainties around these values.
\item  For a practical sample-based solution of the soft HJB equation, we use methods developed in the field of Scientific Machine Learning (SciML) which employs deep neural networks to solve various partial differential equations (PDEs)  from real or simulated data. Such methods are generally known as Neural PDEs and Physics-Informed Neural Networks (PINNs). This paper uses a version of a Neural PDE/PINN  (the \emph{Deep DOCTR-L solver}) that solves the soft HJB equation in a data-driven way by reducing it to a supervised learning problem. 
\item This produces a practical end-to-end decision-making optimizer for a high-dimensional offline policy training from logged 
behavioral data. The way it converts offline off-policy data into an optimal policy is akin to the working of (deep) Q-learning  \cite{Sutton_Barto_2018}. But unlike the latter, the suggested `SciPhy RL' approach is designed for work for high-dimensional continuous state-action spaces, includes \emph{control of uncertainty}, and is `physics-informed' in the sense of its relying on information involving function derivatives (via PDEs) to tame exploration errors of offline RL.  
\end{itemize} 

\subsection{Related work}
\label{sect_Related_work}

\subsubsection{MaxEnt RL and G-learning}

Maximum Entropy (MaxEnt) reinforcement learning (MaxEnt RL) has become one of the most popular approaches to problems of stochastic optimal control (SOC) in a discrete-time setting, see e.g. \cite{Levine_2018} for a review and references to the original literature.
MaxEnt RL focuses on entropy-regularized sample-based approaches to solving Bellman optimality equations for Markov Decision 
Processes. While the latter assume a deterministic policy, MaxEnt RL consider a stochastic (soft) relaxation of classical Bellman equations, where deterministic policies are replaced by \emph{stochastic} policies $ \pi_{\theta} ({\bf a}_t|x_t,t) $ expressed as 
probability distributions (or probability densities, for continuous-action problems) in actions $ {\bf a}_t $ at time $ t $,  parameterized by state variables $ x_t $ and 
model parameters $ \theta $. As any deterministic policy $ a_{\theta} (x_t,t) $ can be thought of as a Dirac-function probability density $ \pi_{\theta}({\bf a}_t |x_t, t)   = 
\delta \left( {\bf a}_t -   a_{\theta} (x_t,t) \right) $, stochastic policies embed all deterministic policies as a special case. 

MaxEnt RL offers a number of attractive new features in comparison to deterministic policies. First, it allows one to quantify uncertainty in suggested optimal actions of the agent. Second, it offers a principled approach to the celebrated exploration-exploitation dilemma \cite{Sutton_Barto_2018}, which usually performs much better than more heuristic algorithms such as $ \varepsilon $-greedy policy randomization schemes.
Third, it makes it possible to learn offline from data collected by another agent whose actions might be sub-optimal. Fourth, MaxEnt RL has a natural counterpart within methods of inverse reinforcement learning (IRL) whose task is to find an unobserved reward function of an agent from a demonstrated behavior, known as MaxEnt IRL. Last but not least, it offers significant computational simplifications by replacing optimization with respect to possible actions by integration, which is often much cheaper computationally.  

Approaches that use stochastic policies require some regularization to prevent infinitely flexible distributions that would overfit data.
 While many MaxEnt RL methods use the Shannon entropy  
 $ \mathcal{H}  \left[ \pi \right]  -  \int   \pi_{\theta}({\bf a}_t |x_t, t) \log  \pi_{\theta}({\bf a}_t |x_t, t) $ as a regularization, 
the G-learning method \cite{G_learning} uses instead a Kullback-Leibler (KL) divergence
 $ KL \left[ \pi || \pi^{0)} \right] $ with some reference (prior) policy $ \pi^{(0)} ({\bf a}_t|x_t,t) $. While this offers a  straightforward generalization of the MaxEnt RL approach
 (as the Shannon entropy is recovered from the KL divergence with a uniform reference density $ \pi^{(0)} $), the G-learning method can be used, in particular, to enforce constraints or prior views on a desired policy $ \pi $.\footnote{For a review of G-learning with either a discrete or continuous state-action space, along with applications in financial modeling, see \cite{MLF}.} In what follows, we will collectively refer to both MaxEnt RL and G-learning as `MaxEnt RL'.
 
\subsubsection{Offline RL}

For many potential applications of RL, an access to a real or simulated environment to try different policies might be too expensive or not feasible.
Offline RL assumes that only a fixed dataset collected under some unknown behavioral policy (or even a number of different behavioral policies, if data are collected from multiple agents) is available to the researcher. 
Developing reliable methods for such an offline RL setting (also referred to as batch-mode RL) have generated considerable interest in the recent literature   \cite{Levine_Offline_RL_2020}. The main challenge with offline RL is that while a fixed dataset may not cover some combinations of states and actions that produce high rewards, a model should rely on some sort of extrapolation in the state-action space. 
With Time-Difference (TD) methods commonly used in RL, rewards are evaluated at actions where there is no data, and propagated through the Bellman equation, potentially bootstrapping errors 
arising due to such extrapolation. 
This is known as the \emph{extrapolation problem} of offline RL \cite{Fujimoto_2019}. 
A number of of approaches focused on constraining policies to not deviate too much from the actual behavioral data were recently proposed in the literature \cite{Fujimoto_2019, Siegel_2020,Levine_Offline_RL_2020}.

 \subsubsection{Reinforcement learning in continuous time}
 
 A continuous-time RL traditionally deals with data-driven methods of solving the classical HJB equation \cite{Doya_2000}. The latter assumes that an optimal policy is deterministic, which is the standard assumption with dynamic programming methods. On the other hand, methods such as MaxEnt RL operate with stochastic policies. Methods based on stochastic policies are often either more computationally efficient or more practically useful (especially for noisy environments) than deterministic policies \cite{Levine_2018}. A soft relaxation of the classical HJB equation that uses stochastic policies in the setting of risk-neutral MaxEnt RL was 
 considered in \cite{Wang_2020}. Practical continuous-time MaxEnt RL methods based on a soft relaxation of HJB equations for deterministic non-linear systems were considered in \cite{Kim_2020}. 
 
 \subsubsection{Distributional RL and risk-sensitive RL}
 
Traditional reinforcement learning approaches focus on policies minimizing the expected value of total return $ Z_T $, as estimated at the current time $ t $, and without controlling for its higher moments. Because higher moments (variance, skewness etc.) control \emph{risk} (uncertainties) of future returns, these methods are sometimes referred to as 'risk-neutral" RL methods.     
Distributional RL  \cite{Bellemare_2017} and risk-sensitive RL \cite{Shen_2014} generalize such 'risk-neutral' methods by incorporating higher-order statistics of $ Z_T $, or even its whole conditional distribution. With risk-sensitive RL, analysis is usually done using a particular risk measure such as e.g. CVAR \cite{Shen_2014},
with a computational approach tuned to this particular risk measure.
On the other hand, Distributional RL proceeds without specifying a particular utility function, but this is achieved at the cost of using policies that only optimize expected values rather than the whole distributions, thus somewhat negating potential advantages offered by a distributional view of learning.
A distributional offline actor-critic method in a discrete-time setting was proposed in \cite{Urpi_2021}.  

\subsubsection{Scientific Machine Learning and Physics-Informed Neural Networks}

Scientific Machine Learning (SciML) is a new sub-field of machine learning research that applies deep neural networks (DNNs) to 
solving partial differential equations (PDEs) and other classical problems of applied mathematics and physics
\cite{SciML}. With approaches developed within SciML, neural networks are used to approximate solutions of PDEs using loss functions that tie up to the structure of the underlying PDE. 
By taking advantage of automatic differentiation, deep learning can provide a mesh-free method and break the curse of dimensionality \cite{Poggio_2017, Grohs_2018}. 

There are a number of advantages in encoding a PDE into a DNN. First, this enforces the PDE as a \emph{constraint} or regularization on data points, which enforces conservation laws for energy, momentum, mass, probability etc., and ensures smoothness of a solution. This produces a very useful '\emph{regularization by a PDE}', as a particular implementation of the main SciML's idea of a `\emph{regularization by a theory}'. 
Second, it reduces the initial infinite-dimensional problem to a finite-dimensional problem in a parameter space. The latter problem can be estimated based on a moderate number of samples, giving rise to sample-efficient schemes. Third,
unlike conventional DNNs that can only be trained on \emph{available} data, PDEs can \emph{predict} the solution of the system for \emph{arbitrary} data. 
This ability is especially valuable for offline RL that faces the extrapolation problem of potentially assigning low costs to state-action combinations not present in the initial data.

Beyond the unifying but general idea of using neural networks to solve PDEs, there are at least three major classes of approaches for solving PDEs with SciML (see \cite{Blechschmidt_2021} and 
\cite{Beck_2020} for a review and references to the original literature). The first approach, called 
Physics-Informed Neural Networks (PINNs), uses a single network as a parameterized approximate solution, and then encodes the original PDE using automatic differentiation as a constraint imposed on a grid of points \cite{PINNs} (see also \cite{DeepXDE} for a brief review along with applications and a TensorFlow-based package for solving PINNs).  Note that the word `physics' used here refers not to potential applications in physical sciences, but rather to the idea that the neural network is provides with information about \emph{derivatives} of functions of interest. It thus can use this information to grasp the notions of 
 smoothness and proximity via the Taylor expansion. The PINNs work well for `tricky' (severely non-linear) low dimensional PDEs 
\cite{PINNs, Blechschmidt_2021} but become inefficient for higher dimensions.     

The other two classes of SciML PDE solvers encode information that goes beyond the PDE itself, and are designed to work better for high-dimensional problems. The second class of methods is based on combining neural networks with the Feynman-Kac formula that expresses a linear backward (Kolmogorov) PDE as a forward expectation amenable to Monte Carlo simulations of the corresponding forward stochastic differential equation (SDE). It uses a neural network to parameterize the solution of the PDE \emph{at a fixed time} for an arbitrary value of its argument \cite{Beck_2018, Beck_2020}. 
 
The third class of SciML methods applies to semilinear, quasi-linear, or fully non-linear PDEs, and is
based on \emph{forward-backward stochastic differential equations} (FBSDEs) to provide computable path-wise approximations to stochastic dynamics associated with a given PDE \cite{Han_2017, Han_2018}. The \emph{Deep BSDE solver} in \cite{Han_2018} uses a neural network to approximate these dynamics. Training 
it on simulated data produces the solution for a fixed state at a fixed point in time - which is similar to the second method, and different from PINNs that offers the solution simultaneously for all times. Other FBSDE methods that produce a solution for arbitrary inputs were also suggested in the literature \cite{Raissi_2018, Zhang_2020}.

\section{Soft HJB equation for distributional reinforcement learning } 
\label{Soft_HJB_control}

\subsection{Probabilistic distributional RL with Kolmogorov PDEs}
\label{sect_cont_time_RL}

Let $ {\bf x}_t \in \mathbb{R}^N $ be a state of the environment at time $ t $, and $ {\bf a}_t \in \mathbb{R}^M $ be an action taken by an agent at that time, 
where $ N $ and $ M $ are dimensions of the state and action spaces, respectively. 
We consider a finite horizon continuous-time control problem on a time interval $ t \in [0, T] $. For an arbitrary intermediate time  $ 0 \leq t \leq T $,  
the total cost $ Z_T $, as seen at time $ t $, is defined as follows\footnote{The factor $ e^{rt} $ in Eq.(\ref{Z_T}) ensures that the total return $ Z_T $ is measured at time $ t $ relative to time $ t = 0 $.} 
\beq
\label{Z_T}
Z_T(t) = \int_{0}^{T} e^{-r (s-t)} c({\bf x}_s, {\bf a}_s) ds
\eeq
where  $ r $ is a continuous-time discount rate and $ c({\bf x}_s, {\bf a}_s) $ is a running cost. Note that the cost  $ c({\bf x}_s, {\bf a}_s) $, as well as other expressions to be presented below, can also explicitly depend on time, but  to ease the notation, time dependences will not be displayed below whenever it does not cause confusion.   

For an arbitrary time $ 0 \leq t < T $, 
$ Z_T(t) $ is a random variable as it depends on future realizations of state-action pairs $ ({\bf x}_s, {\bf a}_s) $ for $ t < s \leq T $. We define the realized cost $ C_t $ as a non-random component of $ Z_T(t) $ which is known at time $  t  $:
\beq
\label{C_t}
C_t =  \int_{0}^{t} e^{-r (s-t)} c({\bf x}_s, {\bf a}_s)ds 
\eeq
so that we have
\beq
\label{Z_T_2}
Z_T(t)  
= C_t + \int_{t}^{T} e^{-r (s-t)} c({\bf x}_s, {\bf a}_s) ds
 \eeq
These definitions imply that at time $ t = T $, the random variable $ Z_T(t) $ becomes observable, with $ Z_T(T) = C_T $. It is therefore convenient to define an extended 
state vector $ {\bf y}_t := ({\bf x}_t, C_t) $, so that the dynamics in $ ({\bf y}_t, {\bf a}_t) $ are Markov. 

The traditional RL is concerned with optimization of the expected total cost  $ J({\bf y}_t)  =  \mathbb{E} \left[ Z_T(t) | {\bf y}_t \right] $. As the expectation is a linear operation, such an objective function is trivially an additive function of $ C_t $ with $ \partial J/ \partial C_t = 1 $. Here we instead 
follow ideas of distributional RL and risk-averse RL, that aim to control the whole distribution of returns rather than only its expected value. Therefore we consider dynamics of a conditional distribution of total return $ Z_T $, as seen at time $ t $ given a state $ {\bf y}_t = ({\bf x}_t, C_t) $:
\beq
\label{P_pi}
P^{\pi} (z | {\bf y}_t, t) := P^{\pi} \left( \left. Z_T(t) = z \right|  {\bf y}_t, t \right)
\eeq
where the superscript $ \pi $ is introduced to emphasize that this conditional probability depends on a policy $ \pi $ (see below). As at time $ t = T $ we have 
$ Z_T(T) = C_T $, this produces a terminal condition on $ P^{\pi} (z | {\bf y}_t, t) $:
\beq
\label{P_pi_T}
P^{\pi} (z | {\bf y}_T, T) = \delta(z - C_T) 
\eeq
Note that unlike the expectation $  \mathbb{E} \left[ Z_T(t) | {\bf y}_t \right] $, the dependence of conditional probability $ P^{\pi} (z | {\bf y}_t, t) $ on $ C_t $ may be non-trivial (in fact, it obeys a backward Kolmogorov equation, see below).

The cost of policy $ \pi $ is determined by the following cost functional:
\beq
\label{J_t} 
J_0^{\pi} ({\bf y}_t,t) = \int dz U(z) P^{\pi} (z | {\bf y}_t, t)  =  \int dz U(z) \int d {\bf a}_t \pi({\bf a}_t| {\bf y}_t) P^{\pi} (z | {\bf y}_t, {\bf a}_t, t)  
\eeq
Here $ U(z)$ is a convex function (a 'negative utility') with a minimum at $ z = 0 $, that measures the amount of undesirability of a total cost $ z $ received on a realized trajectory of state-action pairs on the interval $ t \in [0,T] $. For example, one simple choice is to use $ U(z) = z^2 $, but our approach is general, and enables using an arbitrary function
$ U(x) $ which only impacts a terminal condition, see below. As in MaxEnt RL, we will work with a regularized version of the cost functional
\beq
\label{J_t_G} 
J^{\pi}({\bf y}_t,t) =  J_0^{\pi} ({\bf y}_t,t) +  \frac{1}{\beta} \mathcal{R}^{\pi}({\bf x}_t,t) 
\eeq
where $ \mathcal{R}^{\pi}({\bf x},t)  $ is a time-integrated expected Kullback-Leibler (KL) divergence  
of policy $ \pi $ and a behavioral policy $ \pi_0 $:
\bea
\label{regularizer}
 && \mathcal{R}^{\pi}({\bf x},t) =  
\int_{t}^{T} ds e^{-r (s-t)} \mathbb{E}_{x,t} \left[ \int d {\bf a}_s \pi({\bf a}_s | {\bf x}_s) \log \frac{ \pi({\bf a}_s| {\bf x}_s)}{ \pi_0 ({\bf a}_s | {\bf x}_s) } \right]  
\nonumber \\
 && = 
\int_{t}^{T} ds e^{-r (s-t)} \mathbb{E}_{x,t} \left[ \mathcal{D}_{KL}[ \pi || \pi_0 ]({\bf x}_s,s) \right] 
\eea
The second term in Eq.(\ref{J_t_G}) serves as a regularization that penalizes
large deviations of $ \pi $ from $ \pi_0 $.   
The strength of the regularization is controlled by the `inverse temperature' parameter $ \beta $. While a specific approach to construct a 
behavioral policy $ \pi_0 $ will be presented below, relations derived in this section are general, and apply for any  $ \pi_0 $.

The optimal
cost function $ J ({\bf y}_t,t) $ satisfies the following relation:
\beq
\label{J_t_opt} 
J({\bf y}_t,t) = \min_{\pi} \left\{ J_0^{\pi} ({\bf y}_t,t) +  \frac{1}{\beta} \mathcal{R}^{\pi}({\bf x}_t,t) \right\} 
\eeq
with the terminal condition
\beq
\label{J_T}
J({\bf y}_T,T)  = J^{\pi}({\bf y}_T,T) =  U(C_T) 
\eeq 
Note that while the terminal conditions for the optimal cost functional $ J({\bf y}_T, T) $ and a fixed-policy functional $ J^{\pi}({\bf y}_T,T) $ are both given by the same expression $ U(C_T) $, \emph{realized} values of its argument $ C_T $ depend of the policy $ \pi $. Therefore, in simulations that share the same realizations of random disturbances at every step but vary in sampling actions $ {\bf a}_t \sim \pi(\cdot| x_t) $, realized values of $ C_T $ would in general be different for a fixed policy $ \pi $ and an optimal policy $ \pi = \pi_{\star} $. 

The objective of the agent is to find the optimal policy $ \pi = \pi_{\star} $ by computing the value of optimal cost functional $ J(y_0, 0) $ at time $ t = 0 $. Note that unlike risk-averse RL methods that typically apply a non-linear neg-utility function to each intermediate cost (see e.g.  \cite{Shen_2014}), the convex 'cost price' $ U(z) $ in Eq.(\ref{J_t}) only applies to a single terminal value $ Z_T $. Furthermore, unlike the distributional RL approach that operates on a distributional version of the Bellman equation 
\cite{Bellemare_2017}, we use a probabilistic approach that operates with \emph{probabilities} of random total returns, as seen at different times. With this approach, instead of the traditional RL approach of control \emph{under} uncertainty, here one deals with control \emph{of} uncertainty.  

As the dynamics are Markov in the pair $ ({\bf y}_t, {\bf a}_t) = (x_t, C_t, {\bf a}_t) $,  the conditional probability (\ref{P_pi}) can be expressed in terms of its values at a future time $ s $ such 
that $ t <  s <  T $ by inserting an integral over intermediate states $ {\bf y}_s, {\bf a}_s $: 
\bea
\label{second_eq}
 && P^{\pi} \left( \left. Z_T(t) = z \right| {\bf y}_t, {\bf a}_t, t \right) = 
 \int d{\bf y}_s d {\bf a}_s P^{\pi} ( Z_T(t) = z, {\bf y}_s, {\bf a}_s | {\bf y}_t, {\bf a}_t, t)  \nonumber \\
 && = 
  e^{- r(s - t)}  \int d{\bf y}_s d {\bf a}_s P^{\pi} ( Z_T(s) = z | {\bf y}_s, {\bf a}_s, s, {\bf y}_t, {\bf a}_t, t)  P^{\pi} ( {\bf y}_s, {\bf a}_s, s | {\bf y}_t, {\bf a}_t, t) \\
 &&  = 
  e^{- r(s - t)}  \int d{\bf y}_s d {\bf a}_s P^{\pi} ( Z_T(s) = z | {\bf y}_s, {\bf a}_s, s)  P^{\pi} ( {\bf y}_s, {\bf a}_s, s | {\bf y}_t, {\bf a}_t, t) \nonumber
  \eea
Here the discount factor $ e^{- r(s - t)} $ in the second equation is obtained due to the explicit time dependence on $ t $ in Eq.(\ref{Z_T}).
Plugging this relation into Eq.(\ref{J_t}), we obtain
\bea
\label{J_t_1} 
&& J_0^{\pi} ({\bf y}_t,t)  
= \int dz U(z) \int d {\bf a}_t \int d{\bf y}_s d {\bf a}_s \pi({\bf a}_t| {\bf y}_t) e^{- r(s - t)} P^{\pi} (z | {\bf y}_s, {\bf a}_s,s)  P^{\pi} ( {\bf y}_s, {\bf a}_s, s | {\bf y}_t, {\bf a}_t, t) \nonumber\\  
&& =  e^{- r(s - t)}  \int d{\bf y}_s J^{\pi} ({\bf y}_s,s) \int d {\bf a}_t  \pi({\bf a}_t| {\bf y}_t) P^{\pi} ( {\bf y}_s | {\bf y}_t, {\bf a}_t)  \nonumber \\
&& =   e^{- r(s - t)} \int d{\bf y}_s  J_0^{\pi} ({\bf y}_s,s) P^{\pi} ( {\bf y}_s | {\bf y}_t)   
\eea
Equation (\ref{J_t_1}) relates the cost functional $ J_0^{\pi} ({\bf y}_t,t) $ with its future values at time $ s $. Using Eq.(\ref{J_t_G}), we express it in terms of the 
regularized  cost functional $ J^{\pi} ({\bf y}_t,t) $:
\beq
\label{J_t_pi} 
J^{\pi} ({\bf y}_t,t) - \frac{1}{\beta} \mathcal{R}^{\pi}({\bf x}_t,t) 
 =   e^{- r(s - t)} \int d{\bf y}_s  P^{\pi} ( {\bf y}_s | {\bf y}_t) \left( J^{\pi} ({\bf y}_s,s) -   \frac{1}{\beta} \mathcal{R}^{\pi}({\bf x}_s,s) \right)    
\eeq
To obtain a continuous-time limit, we take $ s = t + \Delta t $ with a small time step $ \Delta t $, 
and expand the expression in the integrand in a Taylor series: 
\bea
\label{J_t_2}
& J^{\pi} ({\bf y}_t,t) - \frac{1}{\beta} \mathcal{R}^{\pi}({\bf x}_t,t) 
 =    e^{- r(s - t)} \int d{\bf y}_s  P^{\pi} ( {\bf y}_s | {\bf y}_t) \left( J^{\pi} ({\bf y}_t,t) - \frac{1}{\beta} \mathcal{R}^{\pi}({\bf x}_t,t) +   \frac{ \partial \left( J^{\pi} -   \frac{1}{\beta} \mathcal{R}^{\pi} \right) }{\partial t} \Delta t  \right. \nonumber \\
 & + 
\frac{\left( J^{\pi} -   \frac{1}{\beta} \mathcal{R}^{\pi} \right)  }{\partial C_t} \Delta C_t 
+  \left. \frac{ \partial \left( J^{\pi} -   \frac{1}{\beta} \mathcal{R}^{\pi} \right) }{\partial {\bf x}_t} \cdot \Delta {\bf x}_t  
+ \frac{1}{2} \frac{ \partial^2 \left( J^{\pi} -   \frac{1}{\beta} \mathcal{R}^{\pi} \right) }{\partial {\bf x}_t^2}  \circ \left( \Delta {\bf x}_t \right)^2  \right) + O \left( \Delta t^2 \right)    
 \eea
where $ (\partial^2 J/ \partial {\bf x}^2) \circ  \Delta {\bf x}^2 = \sum_{i,j} (\partial^2 J/( \partial x_i \partial x_j) \Delta x_i \Delta x_j $,   $ \Delta x_t = x_{t + \Delta t} - x_t $ and $  \Delta C_t = C_{t+\Delta t} - C_t $, and all partial derivatives are computed at time $ t + \Delta t $. The increment $ \Delta C_t $ is obtained from Eq.(\ref{C_t}):
\beq
\label{delta_C_t}
 \Delta C_t =  \left(c (x_t, {\bf a}_t) + r C_t \right)\Delta t + O \left( \Delta t^2 \right)  
\eeq
In addition, we define policy-dependent 'effective' drift, volatility and cost functions by the following relations (here $ s = t + \Delta t  $):
\bea
\label{mu_sigma_pi}
&&  \boldmu ({\bf x}_t, \pi_t) := \lim_{\Delta t \rightarrow 0} \frac{1}{\Delta t} \int d {\bf a}_t  \pi({\bf a}_t | {\bf x}_t) \int d {\bf x}_s P^{\pi} ( {\bf x}_s | {\bf x}_t, {\bf a}_t)   \left( {\bf x}_s - {\bf x}_t \right)
:=  \int d {\bf a}_t  \pi({\bf a}_t | {\bf x}_t)  \boldmu({\bf x}_t, {\bf a}_t)
 \nonumber \\ 
&& \boldsigma^2({\bf x}_t, \pi_t) := \lim_{\Delta t \rightarrow 0} \frac{1}{\Delta t} \int d {\bf a}_t  \pi({\bf a}_t | {\bf x}_t) \int d {\bf x}_s P^{\pi} ( {\bf x}_s | {\bf x}_t, {\bf a}_t)   \left( {\bf x}_s - {\bf x}_t \right) \left( {\bf x}_s - {\bf x}_t \right)^T 
:=  \int d {\bf a}_t  \pi({\bf a}_t | {\bf x}_t)  \boldsigma^2({\bf x}_t, {\bf a}_t)   \nonumber \\
&& c({\bf x}_t, \pi_t) :=  \lim_{\Delta t \rightarrow 0} \int d {\bf a}_t  \pi({\bf a}_t | {\bf x}_t) \int d {\bf x}_s P^{\pi} ( {\bf x}_s | {\bf x}_t, {\bf a}_t) c ({\bf x}_t, {\bf a}_t) 
=   \int d {\bf a}_t  \pi({\bf a}_t | {\bf x}_t)  c ({\bf x}_t, {\bf a}_t)  
\eea 
Here action-dependent drift $ \mu(x_t, {\bf a}_t) $ and volatility $  \sigma({\bf x}_t, {\bf a}_t)  $ 
enter the following controlled stochastic differential equation (SDE):
\beq
\label{CT_Langevin_controlled}
d {\bf x}_t = \boldmu({\bf x}_t, {\bf a}_t) dt + \boldsigma ({\bf x}_t, {\bf a}_t) d {\bf W}_t
\eeq
where $ {\bf W}_t $ is a standard $ N$-dimensional Brownian motion. Parameters $ \boldmu({\bf x}_t, {\bf a}_t) $ and $  \boldsigma({\bf x}_t, {\bf a}_t) $ can be  defined by a model, 
 or alternatively can estimated  from data according to their definitions 
\bea
 \label{mu_sigma_a}
&& \boldmu({\bf x}_t, {\bf a}_t) := \lim_{\Delta t \rightarrow 0} \frac{1}{\Delta t}  \int d {\bf x}_s P^{\pi} ( {\bf x}_s | {\bf x}_t, {\bf a}_t)   \left( {\bf x}_s - {\bf x}_t \right), \; \; \; s = t + \Delta t  
 \nonumber \\ 
&& \boldsigma^2({\bf x}_t, {\bf a}_t) := \lim_{\Delta t \rightarrow 0} \frac{1}{\Delta t}  \int d {\bf x}_s P^{\pi} ( {\bf x}_s | {\bf x}_t, {\bf a}_t)   \left( {\bf x}_s - {\bf x}_t \right)
 \left( {\bf x}_s - {\bf x}_t \right)^T
\eea 
Note that instead of the original SDE (\ref{CT_Langevin_controlled}) where $ {\bf a}_t $ serves as a random parameter, the dynamics in the state space only
underlying Eq.(\ref{J_t_2}) involves taking the expectation over $ {\bf a}_t $ with the policy distribution $ \pi $:
\beq
\label{CT_Langevin_controlled_pi}
d {\bf x}_t = \boldmu({\bf x}_t, \pi_t) dt + \boldsigma ({\bf x}_t, \pi_t) d {\bf W}_t
\eeq
Using relations (\ref{delta_C_t})-(\ref{CT_Langevin_controlled_pi}), taking the continuous time limit $ \Delta t = dt \rightarrow 0 $ in Eq.(\ref{J_t_2}), and using the backward (Kolmogorov) equation for the 
expected KL-divergence $  \mathbb{E}_{x,t} \left[ \mathcal{D}_{KL}[ \pi || \pi_0 ]({\bf x}_s,s) \right] $, we obtain the backward PDE for the cost functional
$ J^{\pi} $:
\beq
\label{Kolmogorov_J}
- \frac{\partial J^{\pi}}{\partial t} =    \left(c ({\bf x}_t, \pi_t) + r C_t \right) \frac{ \partial J^{\pi}}{\partial C_t} + \boldmu({\bf x}_t, \pi_t) \cdot \frac{ \partial J^{\pi}}{\partial {\bf x}_t} + 
  \frac{1}{2} \boldsigma^2({\bf x}_t, \pi_t) \circ \frac{ \partial^2 J^{\pi}}{\partial {\bf x}_t^2} - r J^{\pi} + \frac{1}{\beta}  \mathcal{D}_{KL}[ \pi || \pi_0 ]({\bf x}_t,t) 
 \eeq  
 which should be supplemented by the terminal condition (\ref{J_T}). The conditional probability $ P^{\pi}(z|{\bf y}_t, t) $ satisfies a similar backward Kolmogorov PDE, but 
 without the last term:
 \beq
 \label{Kolmogorov_P}
- \frac{\partial P^{\pi}}{\partial t} =   \left(c ({\bf x}_t, \pi_t) + r C_t \right)  \frac{ \partial P^{\pi}}{\partial C_t} + \boldmu({\bf x}_t, \pi_t) \cdot \frac{ \partial P^{\pi}}{\partial {\bf x}_t} + 
  \frac{1}{2} \boldsigma^2({\bf x}_t, \pi_t) \circ \frac{ \partial^2 P^{\pi}}{\partial {\bf x}_t^2} - r  P^{\pi}
 \eeq  
 Given a policy $ \pi $, Eq.(\ref{Kolmogorov_P}) can be solved backward in time starting with $ t = T $ using the terminal condition (\ref{P_pi_T}),  
 to produce the time-0 conditional distribution $ P^{\pi} (z | {\bf y}_0, 0) $ of the cumulative return $ Z_T $. The latter distribution  $ P^{\pi} (z | {\bf y}_0, 0) $ quantifies the uncertainty about the value of  $ Z_T $ as seen at time $ t = 0 $. The quality of control obtained
 using an optimal policy $ \pi = \pi_{\star} $, instead of a given policy $ \pi $, can be judged by comparing how both policies $ \pi $ and $ \pi_{\star} $ impact a solution of the backward PDE (\ref{Kolmogorov_P}) at $ t = 0 $. 
 
\subsection{Soft HJB equation}
\label{sect_Soft_HJB_equation} 
 
To find the optimal policy $ \pi = \pi_{\star} $, we need to solve 
an equation for the optimal cost functional $ J({\bf y},t) = J^{\star}({\bf y},t) $, which is obtained from Eq.(\ref{Kolmogorov_J}) by taking a minimum over 
all policies $ \pi $:
\bea
\label{Kolmogorov_J_opt}
- \frac{\partial J}{\partial t} + r J -  r C_t  \frac{ \partial J}{\partial C_t} 
&=&  \min_{\pi} \int d {\bf a}_t \pi({\bf a}_t| {\bf x}_t) 
\left[   c ({\bf x}_t, {\bf a}_t) \frac{ \partial J}{\partial C_t} + \boldmu({\bf x}_t, {\bf a}_t)  \cdot \frac{ \partial J}{\partial {\bf x}_t} \right. \nonumber \\
&& \left. + 
  \frac{1}{2} \boldsigma^2({\bf x}_t, {\bf a}_t)  \circ \frac{ \partial^2 J}{\partial {\bf x}_t^2} + \frac{1}{\beta} \frac{ \pi({\bf a}_t| {\bf x}_t)}{ \pi_0({\bf a}_t|{\bf x}_t)} \right]  
 \eea  
The minimum over polices $ \pi $ in Eq.(\ref{Kolmogorov_J_opt}) can be computed analytically in terms of $ J $ and its derivatives:
\beq
\label{pi_opt}
\pi({\bf a}_t| {\bf x}_t) = \frac{1}{Z(J,{\bf x}_t,t) } \pi_0({\bf a}_t| {\bf x}_t) \exp{ \left[ - \beta \left( c ({\bf x}_t, {\bf a}_t)  \frac{\partial J}{ \partial C_t} +  \boldmu({\bf x}_t, {\bf a}_t) \cdot \frac{ \partial J}{\partial {\bf x}_t} + 
  \frac{1}{2} \boldsigma^2({\bf x}_t, {\bf a}_t)  \circ \frac{ \partial^2 J}{\partial {\bf x}_t^2} \right) \right] }
\eeq
where $ Z(J,{\bf x}_t,t) $ is a normalization factor:
\beq
\label{Z_t}
Z(J,{\bf x}_t,t) = \int d {\bf a}_t   \pi_0({\bf a}_t| {\bf x}_t)  \exp{ \left[ - \beta \left( c ({\bf x}_t, {\bf a}_t)  \frac{\partial J}{ \partial C_t} +  \boldmu({\bf x}_t, {\bf a}_t)  \cdot \frac{ \partial J}{\partial {\bf x}_t} + 
  \frac{1}{2} \boldsigma^2({\bf x}_t, {\bf a}_t) \circ \frac{ \partial^2 J}{\partial {\bf x}_t^2} \right) \right] }
\eeq
Plugging the optimal policy (\ref{pi_opt}) back into Eq.(\ref{Kolmogorov_J_opt}), we obtain:
\bea
\label{Kolmogorov_J_opt_nonlin}
- \frac{\partial J}{\partial t}  
&=& - \frac{1}{\beta} \log  \int d {\bf a}_t   \pi_0({\bf a}_t| {\bf x}_t)  \exp{ \left[ - \beta \left( c ({\bf x}_t, {\bf a}_t)  \frac{\partial J}{ \partial C_t} + 
 \boldmu({\bf x}_t, {\bf a}_t)  \cdot \frac{ \partial J}{\partial {\bf x}_t} + 
  \frac{1}{2} \boldsigma^2({\bf x}_t, {\bf a}_t) \circ \frac{ \partial^2 J}{\partial {\bf x}_t^2} \right) \right] } \nonumber \\
 & + &  r C_t  \frac{ \partial J}{\partial C_t} - r J 
 \eea
 To obtain a more tractable form, we fix the dependence on actions $ {\bf a}_t $ in parameters $   c ({\bf x}_t, {\bf a}_t)$,  $ \mu({\bf x}_t, {\bf a}_t) $ and $  \sigma^2({\bf x}_t, {\bf a}_t) $ as follows:
 \beq
 \label{lin_quadratic}
  c ({\bf x}_t, {\bf a}_t) =  c_0 ({\bf x}_t) + \frac{1}{2}  c_1 ({\bf x}_t) || {\bf a}_t ||^2, \; \;  \;  \boldmu({\bf x}_t, {\bf a}_t) =   \boldmu_0({\bf x}_t) + \boldmu_1({\bf x}_t) {\bf a}_t, \; \; \;   \boldsigma^2({\bf x}_t, {\bf a}_t)  =  \boldsigma^2({\bf x}_t)
 \eeq 
 where functions $ c_0 ({\bf x}_t), c_1 ({\bf x}_t) \geq 0,  \, \forall  {\bf x}_t $ to ensure that the running cost $  c ({\bf x}_t, {\bf a}_t) $ is non-negative.\footnote{
 The particular linear-quadratic dependence on $ {\bf a}_t $ in Eq.(\ref{lin_quadratic}) can be interpreted as leading-order Taylor expansions of 
 more general functions $   c ({\bf x}_t, {\bf a}_t)  $ and $  \boldmu({\bf x}_t, {\bf a}_t) $.
 Note that while the dependence on $ {\bf a}_t $ in Eq.(\ref{lin_quadratic}) coincides with the standard choice in the setting of a deterministic control with the (standard) HJB equation, here we deal with \emph{stochastic} policies. The latter can produce probabilistic scenarios capturing non-quadratic effects in the costs or dynamics even with a linear-quadratic dependence on \emph{realized} values of actions $ {\bf a}_t $.
 Therefore, using stochastic policy, the linear-quadratic specification in Eq.(\ref{lin_quadratic}) still retains flexibility to capture non-quadratic effects, which is unlike the case with deterministic policies where it leads to neglecting \emph{all} non-quadratic effects.}
   
 With these specifications, Eq.(\ref{pi_opt}) simplifies as follows:
\beq
\label{pi_opt_2}
\pi({\bf a}_t| {\bf x}_t) = \frac{1}{Z(J,{\bf x}_t,t) } \pi_0({\bf a}_t| {\bf x}_t) \exp{ \left[ - \beta \left(  \frac{1}{2} c_1 ({\bf x}_t) || {\bf a}_t ||^2 \frac{\partial J}{ \partial C_t} 
+  {\bf a}_t^T \boldmu_1^T({\bf x}_t) \cdot \frac{ \partial J}{\partial x_t} \right) \right] }
\eeq
where 
\beq
\label{Z_t_2}
Z(J,{\bf x}_t,t) = \int d {\bf a}_t   \pi_0({\bf a}_t| {\bf x}_t)  \exp{ \left[ - \beta \left(  \frac{1}{2} c_1 ({\bf x}_t) || {\bf a}_t ||^2 \frac{\partial J}{ \partial C_t} 
+  {\bf a}_t^T \boldmu_1^T({\bf x}_t) \cdot \frac{ \partial J}{\partial x_t}  \right)  \right] }
\eeq
Plugging these relations into Eq.(\ref{Kolmogorov_J_opt_nonlin}), we obtain
\bea
\label{Kolmogorov_J_opt_nonlin_2}
- \frac{\partial J}{\partial t}  
&=&  \left( c_0({\bf x}_t) + r C_t \right)  \frac{ \partial J}{\partial C_t} +  \boldmu_0 ({\bf x}_t)  \cdot \frac{ \partial J}{\partial {\bf x}_t} + 
  \frac{1}{2} \boldsigma^2({\bf x}_t) \circ \frac{ \partial^2 J}{\partial {\bf x}_t^2}- r J \nonumber \\
& - & \frac{1}{\beta} \log  \int d {\bf a}_t   \pi_0({\bf a}_t| {\bf x}_t)  \exp{ \left[ - \beta \left( \frac{1}{2} c_1 ({\bf x}_t) || {\bf a}_t ||^2 \frac{\partial J}{ \partial C_t} + 
  {\bf a}_t^T \boldmu_1^T({\bf x}_t)  \cdot \frac{ \partial J}{\partial {\bf x}_t} \right) \right] }    
 \eea
 To further simplify the resulting equation, we use a Gaussian mixture (GM) policy as a model of the behavioral policy $ \pi_0({\bf a}_t | {\bf x}_t) $:
 \beq
 \label{pi_0_GM}
 \pi_0({\bf a}_t | {\bf x}_t) = \sum_{k=1}^{K} \omega_k \mathcal{N} \left( {\bf a}_t | {\bf u}_k({\bf x}_t,t), \sigma_k \mathbb{I} \right), \; \;  \sum_{k=1}^{K} \omega_k  = 1, \; \; \; 0 \leq \omega_k \leq 1  
 \eeq
 Here the Gaussian means $  {\bf u}_k = {\bf u}_k({\bf x}_t,t) $ can be simple linear functions of $ {\bf x}_t $. Alternatively, non-linear specifications could be considered using e.g. neural networks.
 Covariance matrices  $ \sigma_k \mathbb{I} $ specified by scalar parameters $ \sigma_k $ are isotropic and state-independent.
 Gaussian weights $ \omega_k $ are assumed to be independent of the state  $ {\bf x}_t $ and time $ t $, but may be made dependent on them if needed without added computational complexity. Note that the quadratic dependence on $ {\bf a}_t $ in Eq.(\ref{pi_opt_2}) implies that the optimal policy $ \pi $ is also given by a Gaussian mixture with different weights, means and variances that now all depend on partial derivatives of the cost function:
 \beq
 \label{pi_GM}
 \pi({\bf a}_t | {\bf x}_t) = \sum_{k=1}^{K} \omega_k \left[ J \right] \mathcal{N} \left( \left. {\bf a}_t  \right| 
{\bf u}_k \left[J \right],   \Omega_k\left[J \right] \right)
 \eeq
 where 
 \bea
 \label{means_vars}
 && \omega_k \left[J \right]  = \frac{ \omega_k e^{ - \beta \mathcal{H}_k \left[J \right]}}{ \sum_{k}  \omega_k e^{ - \beta \mathcal{H}_k \left[J \right]}} 
\nonumber  \\
&& {\bf u}_k \left[J \right] = \frac{{\bf u}_k({\bf x}_t,t) - \beta \sigma_k^2 \boldmu_1({\bf x}_t) \cdot
 \frac{\partial J}{\partial {\bf x}_t}}{ 1 + \beta \sigma_k^2 c_1 ({\bf x}_t) \frac{\partial J}{ \partial C_t}}, \; \; \; \Omega_k\left[J \right] = \left( \frac{1}{\sigma_k^2} + \beta  c_1 ({\bf x}_t) \frac{\partial J}{ \partial C_t}
 \right)^{-1} \mathbb{I}
 \eea
 and Hamiltonians $ \mathcal{H}_k \left[J \right] $ are the following non-linear functionals of $ \partial J/ \partial {\bf x}_t $ and $ \partial J/ \partial C_t $:
\beq
\label{H_J}
\mathcal{H}_k \left[J \right] =  \frac{1}{2} \frac{ {\bf u}_k^T {\bf u}_k 
c_1 ({\bf x}_t) \frac{\partial J}{\partial C_t} + 2 {\bf u}_k^T \boldmu_1  \cdot \frac{ \partial J}{\partial {\bf x}_t}  - 
 \beta \sigma_k^2  
\left(\boldmu_1 \cdot \frac{ \partial J}{\partial {\bf x}_t} \right)^2 
}{ 1 +  \beta \sigma_k^2 c_1 ({\bf x}_t) \frac{\partial J}{\partial C_t} } 
+ \frac{1}{2 \beta} \log \left( 1 +  \beta \sigma_k^2 c_1 ({\bf x}_t) \frac{\partial J}{ \partial C_t} \right)
 \eeq
 We refer to $ \mathcal{H}_k \left[J \right] $ as Hamiltonians because they serve as energies when the updated weights $  \omega_k \left[J \right] $ are seen as a Boltzmann distribution with  $ \mathcal{H}_k \left[J \right]$ being the energy of the $ k$-th state where $ k = 1, \ldots, K $ and $ K $ is the number of Gaussian of components in the Gaussian mixture.
 
Eq.(\ref{pi_GM}) offers an interpretation of policy optimization as re-weighing of Gaussian components according to energies $  \mathcal{H}_k \left[J \right] $,
and also adjusting their means and variances according to Eqs.(\ref{means_vars}). This means that by using Gaussian mixtures to model \emph{behavioral} policies,
the present formalism enables modeling potentially multi-modal \emph{optimal} policies using flexible transformations of behavioral policies which stay within the class of Gaussian mixtures. 

 As is clearly seen from Eq.(\ref{pi_GM}), in the high-temperature limit $ \beta \rightarrow 0 $, we obtain $ \pi({\bf a}_t | {\bf x}_t)  = \pi_0({\bf a}_t | {\bf x}_t)  $, meaning the absence of policy optimization in this limit. On the other hand, in the low-temperature limit $ \beta \rightarrow \infty $, we obtain a deterministic policy
with zero volatility and the fixed value
\beq
\label{a_deterministic}
{\bf a}_t = - \frac{\boldmu_1({\bf x}_t)}{c_1 ({\bf x}_t)}  \cdot \frac{ \partial J / \partial {\bf x}_t}{ \partial J / \partial C_t} \; \; \; \; (\beta \rightarrow \infty)
\eeq  
Using the GM prior policy (\ref{pi_0_GM}), the integral in Eq.(\ref{Kolmogorov_J_opt_nonlin_2}) can be computed using the following formula valid for an 
arbitrary matrix $ {\bf C} $ and vector $ {\bf D} $:
\beq
\label{Gaussian_integral_exp}
\frac{\int d {\bf x} e^{ - \frac{1}{2} ({\bf x} - \bar{\bf x})^T  {\bf \Sigma}^{-1} ({\bf x} - \bar{\bf x})
 - \frac{1}{2} {\bf x}^T  {\bf C} {\bf x}  -  {\bf x}^T  {\bf D}} }{(2 \pi)^{\frac{N}{2}} \left| {\bf \Sigma} \right|^{\frac{1}{2}}} = 
 \frac{ e^{ \frac{1}{2} \left( {\bf C} \bar{\bf x} + {\bf D} \right)^T \left( {\bf \Sigma}^{-1} + {\bf C} \right)^{-1} \left( {\bf C} \bar{\bf x} + {\bf D} \right)
 -  \frac{1}{2} \bar{\bf x}^T {\bf C} \bar{\bf x} -  \bar{\bf x}^T {\bf D}} }{
 \left| {\bf \Sigma} \right|^{ \frac{1}{2}}   \left| {\bf \Sigma}^{-1} + {\bf C} \right|^{ \frac{1}{2}} } 
 \eeq
 Using this relation in Eq.(\ref{Kolmogorov_J_opt_nonlin_2}), we obtain
\beq
\label{Kolmogorov_J_opt_nonlin_GM}
 - \frac{\partial J}{\partial t}  
=  \left( c_0({\bf x}_t)  + r C_t \right)  \frac{ \partial J}{\partial C_t} + 
 \boldmu_0 ({\bf x}_t) 
\cdot \frac{ \partial J}{\partial {\bf x}_t} + 
  \frac{1}{2} \boldsigma^2({\bf x}_t) \circ \frac{ \partial^2 J}{\partial {\bf x}_t^2}- r J 
 - \frac{1}{\beta} \log \sum_{k} \omega_k e^{ - \beta \mathcal{H}_k \left[ J \right] }
\eeq
The semilinear backward PDE  (\ref{Kolmogorov_J_opt_nonlin_GM})  (the  'soft HJB equation'), or its more general form in  (\ref{Kolmogorov_J_opt_nonlin}), is the main theoretical result of this paper. It can be viewed as  a probabilistic relaxation of the classical HJB equation for distributional learning that aims at control of the whole return distribution rather than only the expected returns. The resulting PDEs (\ref{Kolmogorov_J_opt_nonlin}) and  (\ref{Kolmogorov_J_opt_nonlin_GM}) are therefore  
different from a PDE obtained in \cite{Wang_2020} which addressed a more traditional `risk-neutral' RL in the continuous time formulation. 
The formalism developed in the present paper, by extending the state space and including additional partial derivatives $ \sim \partial J / \partial C_t $, enables extending this approach to a risk-sensitive setting, with a small computational overhead. 

The classical HJB equation is recovered from  Eq.(\ref{Kolmogorov_J_opt_nonlin_GM}) in the `zero-temperature' limit $ \beta \rightarrow \infty $, where the Hamiltonian (\ref{H_J}) has a simpler form
\beq
\label{H_J_2}
\mathcal{H}_k \left[ J \right] = - \frac{1}{2} \frac{ \left( \boldmu_1 \cdot \frac{ \partial J}{\partial {\bf x}_t} \right)^2}{ c_1 ({\bf x}_t) \frac{\partial J}{\partial C_t}  } \; \; \; (\beta \rightarrow \infty) 
\eeq
Note that in this limit $ \mathcal{H}_k \left[ J \right] $ becomes independent of $ k $, and therefore the dependence on the prior weights $ \omega_k $ 
drops from the problem. 
Using this expression, the zero-temperature limit of Eq.(\ref{Kolmogorov_J_opt_nonlin_GM}) reads
\beq
\label{Kolmogorov_J_opt_nonlin_GM_zero_temp}
 - \frac{\partial J}{\partial t}  
=  \left( c_0({\bf x}_t)  + r C_t \right)  \frac{ \partial J}{\partial C_t} + 
 \boldmu_0 ({\bf x}_t) 
\cdot \frac{ \partial J}{\partial {\bf x}_t} + 
  \frac{1}{2} \boldsigma^2({\bf x}_t) \circ \frac{ \partial^2 J}{\partial {\bf x}_t^2} - r J 
 - \frac{1}{2}   \frac{ \left(  \boldmu_1({\bf x}_t) \cdot \frac{ \partial J}{\partial {\bf x}_t} \right)^2 }{ c_1 ({\bf x}_t) \frac{ \partial J}{\partial C_t}} 
 \eeq
Therefore, the zero-temperature limit $ \beta \rightarrow \infty $ of the soft HJB equation (\ref{Kolmogorov_J_opt_nonlin_GM}) reproduces the 
classical HJB equation.\footnote{ 
As was remarked in Sect.~\ref{sect_cont_time_RL}, with the conventional `risk-neutral' optimal control that minimizes the conditional expectation $ \mathbb{E} \left[ \left. {\bf Z}_t \right| {\bf x}_t \right] $, we 
would trivially have $  \frac{ \partial J}{\partial C_t} = 1 $. In this case, Eq.(\ref{Kolmogorov_J_opt_nonlin_GM_zero_temp}) corresponds to a more familiar form of the HJB equation.}  
It also recovers a deterministic-policy optimization of the classical HJB equation.  
In this limit, optimization w.r.t. stochastic policies becomes an unconstrained optimization w.r.t. all probability distributions, and is therefore equivalent to a point-wise optimization w.r.t. all actions $ {\bf a}_t $, producing Eq.(\ref{a_deterministic}).

On the other hand, in a 'high-temperature' limit  $ \beta \rightarrow 0 $, the Hamiltonians $ \mathcal{H}_k \left[ J \right] $ have the following form:
\bea
\label{H_J_high_T}
&& \mathcal{H}_k \left[ J \right] =  \frac{1}{2}  \left( {\bf u}_k^T {\bf u}_k  + \sigma_k^2 \right) c_1 ({\bf x}_t)
\frac{\partial J}{\partial C_t} +  {\bf u}_k^T \boldmu_1 ({\bf x}_t) \cdot \frac{ \partial J}{\partial {\bf x}_t } \\
&& + \beta  \sigma_k^2  \left(    \frac{1}{2} \left( {\bf u}_k^T {\bf u}_k  + \sigma_k^2 \right) c_1^2({\bf x}_t) 
\left(\frac{\partial J}{\partial C_t} \right)^2  +   
+  {\bf u}_k^T \boldmu_1 ({\bf x}_t) \cdot \frac{ \partial J}{\partial {\bf x}_t } c_1 ({\bf x}_t) \frac{\partial J}{\partial C_t}   
- \frac{1}{2}   \left( \boldmu_1 \cdot \frac{ \partial J}{\partial {\bf x}_t} \right)^2   \right)  + O( \beta^2) \nonumber 
 \eea
In the strict limit $ \beta = 0 $, the cost of information update
from the prior policy  $ \pi^{(0)} $ becomes prohibitively high, and the agent proceeds with the prior policy $ \pi^{(0)} $ without trying to further optimize it. 
The optimality equation (\ref{Kolmogorov_J_opt_nonlin}) in this limit coincides with the linear equation 
(\ref{Kolmogorov_J}) where we should substitute $ \pi = \pi_0 $. On the other hand, sub-leading terms $ O(\beta) $ in Eq.(\ref{H_J_high_T}) could be used to construct corrections in a high-temperature limit of the soft HJB equation  (\ref{Kolmogorov_J_opt_nonlin_GM}).

\subsection{Path probabilities under different drifts}
\label{sect_path_probs}

As we saw above, the `effective' diffusion process in the $ {\bf X} $ space relevant for solving the policy optimization process involves taking expectations with respect to actions, and  
is given by Eq.(\ref{CT_Langevin_controlled_pi}) which we repeat here
\beq
\label{CT_Langevin_controlled_pi_2}
d {\bf x}_t = \boldmu({\bf x}_t, \pi_t) dt + \boldsigma ({\bf x}_t, \pi_t) d {\bf W}_t
\eeq
With our specifications in Eqs.(\ref{lin_quadratic}), we have
\beq
\label{drift_sigma_pi}
 \boldmu({\bf x}_t, \pi_t)  = \boldmu_0({\bf x}_t) + \boldmu_1({\bf x}_t) \langle {\bf a}_t \rangle_{\pi}, \; \; \; 
  \boldsigma ({\bf x}_t, \pi_t) =  \boldsigma ({\bf x}_t)
  \eeq
  where $ \langle {\bf a}_t \rangle_{\pi} $ is the expected value of $ {\bf a}_t $ under policy $ \pi $.
  
 In simulation-based settings, Eq.(\ref{CT_Langevin_controlled_pi_2}) is typically used by simulating trajectories of the Brownian motion $ {\bf W}_t $, which are then used to forward-propagate the state variable $ {\bf x}_t $ starting with an initial value $ {\bf x}_0 $ at time $ t = 0 $. 
 In the present setting, we deal with offline learning where instead of a set of Brownian trajectories $ {\bf W}_t $, 
 we are directly given a fixed set of trajectories of $ {\bf x}_t $. Therefore, it is convenient to switch from a Brownian path measure $ \mathcal{D} W_t $ to a path integral measure $ \mathcal{D} X_t $.    

This can be done starting with a time-discretized version of Eq.(\ref{CT_Langevin_controlled_pi_2}), with time steps $ \Delta t $.
The short time (with $ \Delta t \rightarrow 0 $) transition probabilities can be expressed as functions of the state $ {\bf x}_t $ if we 
take the $ N $ dimensional Gaussian distribution of $ \Delta {\bf W}_t $, and replace the values $ \Delta {\bf W}_t $ in this equation using the discrete version of Eq.(\ref{CT_Langevin_controlled_pi_2}). This produces the well-known transition probability of a multi-dimensional diffusion process
\beq
\label{prob_trans_path}
P( {\bf x}_{t+\Delta t} | {\bf x}_t) = \frac{1}{ \sqrt{ \left( 2 \pi \Delta t \right)^{N}  | \boldsigma \boldsigma^T |}} e^{ - S({\bf x}_t, {\bf x}_{t+\Delta t}, \mu)}
\eeq
where
\beq
\label{action}
 S({\bf x}_t, {\bf x}_{t+\Delta t}, \mu) = \frac{\Delta t}{2} \sum_{i,j} \left[ \boldsigma \boldsigma^T \right]_{ij} ^{-1} \left( \xdot_i 
 -  \boldmu_i({\bf x}_t, \pi_t)
 \right)  \left( \xdot_j -  \boldmu_j({\bf x}_t, \pi_t)  \right) := \mathcal{L} ({\bf x}_t, \xdot_t, \mu) \Delta t  
\eeq
( here $ \xdot_i = dx_i/ dt $ stands for the time derivative) is the action on the trajectory $ ({\bf x}_t, {\bf x}_{t+\Delta t}) $, and 
$  \mathcal{L} ({\bf x}_t, \xdot_t, \mu) $ is the Lagrangian.\footnote{The action shown in Eq.(\ref{action}) corresponds to It{\^o}'s definition of a discretization scheme for the continuous-time SDE (\ref{CT_Langevin_controlled_pi_2}).} A similar relation to (\ref{prob_trans_path}) can be used for a finite time interval $ [0, T] $. In this 
case, the short-term action exponent $ S({\bf x}_t, {\bf x}_{t+\Delta t}, \mu)  = \mathcal{L} ({\bf x}_t, \xdot_t, \mu) \Delta t  $ is replaced by the integral for the total action $ \int_{0}^{T}  \mathcal{L} ({\bf x}_t, \xdot_t, \mu) d t $. Such continuous-time limit produces a path integral formulation of a multi-dimensional diffusion process, 
see e.g. \cite{Bennati_1999}. 

The transition probability formula (\ref{prob_trans_path}) can now be used to obtain the likelihood ratio of 
a given transition $ {\bf x}_t \rightarrow {\bf x}_{t + \Delta t} $ under two different drift functions $ \boldmu^{(0)}({\bf x}_t, t)  $ and  $ \boldmu^{(1)}({\bf x}_t, t)  $.
The likelihood ratio is 
\beq
\label{likelihood_ratio}
\frac{ P^{(\mu^{(1)})} ( {\bf x}_{t+\Delta t} | {\bf x}_t)}{  P^{(\mu^{(0)})} ( {\bf x}_{t+\Delta t} | {\bf x}_t)}  = e^{ - \left( S({\bf x}_t, {\bf x}_{t+\Delta t}, \mu_1) - 
 S({\bf x}_t, {\bf x}_{t+\Delta t}, \mu_0) \right) } = e^{ - \Delta  S({\bf x}_t, {\bf x}_{t+\Delta t}) }
 \eeq
where 
\beq
\label{delta_S}
\Delta S({\bf x}_t, {\bf x}_{t+\Delta t}) =   
\frac{\Delta t}{2} \sum_{i,j} \left[ \boldsigma \boldsigma^T \right]_{ij} ^{-1} \left( \boldmu_i^{(1)} \boldmu_j^{(1)} - 
\boldmu_i^{(0)} \boldmu_j^{(0)} 
- \xdot_i \left( \boldmu_j^{(1)} -  \boldmu_j^{(0)}  \right) -  \xdot_j \left( \boldmu_i^{(1)} -  \boldmu_i^{(0)}  \right) \right)
\eeq
The last relation (\ref{likelihood_ratio}) is very convenient in our setting, as it gives the likelihood ratio for a given transition at $ [t, t + \Delta t] $ under drifts induced by policies $ \pi $ and $ \pi_0 $ in terms of state variables directly observed at these times. In a data-driven setting of offline learning, 
this method is more convenient than using the Girsanov theorem that expresses likelihood ratios such as (\ref{likelihood_ratio}) in terms of integrals of the Brownian motion \cite{Karatzas_Shreve}, though it is equivalent to the latter. To see this, consider a 1D case for Eq.(\ref{delta_S}) with $ \boldmu^{(1)}  = 
\mu(x_t,t) $ and $ \boldmu^{(0)}  = 0 $. In the infinitesimal limit $ \Delta t = dt \rightarrow 0 $, we obtain
\beq
\label{delta_S_1D}
\Delta S =   
\frac{1}{\sigma^2} \left( \frac{1}{2} \mu^2(x_t,t)  dt  
-  \mu(x_t,t)  d  x_t \right) = \frac{1}{\sigma^2} \left( - \frac{1}{2} \mu^2(x_t,t)  dt  
-  \mu(x_t,t) \sigma d W_t \right)
\eeq
where in the last step we used Eq.(\ref{CT_Langevin_controlled_pi_2}). Using this expression in (\ref{likelihood_ratio}) gives the Girsanov theorem representation for the likelihood ratio  \cite{Karatzas_Shreve}.

\subsection{Forward-Backward SDEs and the Hamilton-Jacobi equation}
\label{sect_FBSDE}

Semilinear PDEs such as Eq.(\ref{Kolmogorov_J_opt_nonlin_GM}) can be explored using stochastic dynamics corresponding to such PDEs. These dynamics are given in terms of a coupled pair of a forward and backward SDEs. 
The forward SDE is given by Eq.(\ref{CT_Langevin_controlled_pi_2}) which we repeat here:
\beq
\label{CT_Langevin_controlled_pi_3}
d {\bf x}_t =  \left(  \boldmu_0({\bf x}_t) + \boldmu_1({\bf x}_t) \langle {\bf a}_t \rangle \left[ J \right] \right) dt + \boldsigma ({\bf x}_t) d {\bf W}_t
\eeq
where we now write $  \langle {\bf a}_t \rangle \left[ J \right] $ instead of $  \langle {\bf a}_t \rangle_{\pi} $ to emphasize that the expected action 
$ \langle {\bf a}_t \rangle_{\pi} $ is now viewed as a functional of derivatives of the cost function $ J({\bf x}_t,t ) $.
The expected action $  \langle {\bf a}_t \rangle \left[ J \right] $ that enters this equation can be read off Eq.(\ref{pi_GM}):
\beq
\label{a_mean_pi}
 \langle {\bf a}_t \rangle \left[ J \right]  = 
  \sum_{k=1}^{K} \omega_k\left[ J \right]  \frac{{\bf u}_k({\bf x}_t,t) - \beta \sigma_k^2 \boldmu_1({\bf x}_t) 
 \frac{\partial J}{\partial {\bf x}_t}}{ 1 + \beta \sigma_k^2c_1 ({\bf x}_t)  \frac{\partial J}{ \partial C_t}} 
 \eeq
A second SDE is obtained by combining It{\^{o}}'s lemma for $ J({\bf x}_t, t) $ with the backward PDE (\ref{Kolmogorov_J_opt_nonlin_GM}). This produces 
\beq
\label{backward_SDE}
d J = \left( r J + \frac{1}{\beta} \log \sum_{k} \omega_k e^{ - \beta \mathcal{H}_k\left[ J \right] } \right) dt + \frac{\partial J}{\partial {\bf x}_t} \boldsigma({\bf x}_t) d {\bf W}_t  
\eeq
with the terminal condition $ J({\bf x}_T, C_T,T) = U(C_T) $ (see Eq.(\ref{J_T})). As it needs to be solved starting with $t = T $, this equation is referred to as the backward SDE.

Coupled systems of forward-backward SDEs (FBSDEs) such as our Eqs.(\ref{CT_Langevin_controlled_pi_3}, \ref{backward_SDE}) are most commonly used 
for semilinear PDEs such as Eq.(\ref{Kolmogorov_J_opt_nonlin_GM}) in a model- and simulation-based Monte Carlo setting. With these methods, 
simulated paths of a Brownian motion $ {\bf W}_t $ are first used for 
obtain forward paths of $ {\bf x}_t $ on $ t \in [0, T] $ starting with $ t = 0 $, and then are used again in reverse on the backward path, to find a time-0 value of $ J $ by backward 
recursion of Eq.(\ref{backward_SDE}) starting with the terminal value $ J(T) =  U(C_T) $.

Unlike such a simulation-based approach, here we are concerned with \emph{offline learning}, where instead of observing paths of a Brownian motion $ {\bf W}_t $, we have a fixed set of observed (realized) trajectories $ {\bf x}_t $. We may think of these trajectories as realizations of an unobserved 
Brownian motion $ {\bf W}_t $. The latter should not be known explicitly, as we can  
 use Eq.(\ref{CT_Langevin_controlled_pi_3}) to eliminate  $ {\bf W}_t $ 
from Eq.(\ref{backward_SDE}):
\beq
\label{backward_X}
d J  =    \left( r J - \left( \boldmu_0({\bf x}_t) + \boldmu_1({\bf x}_t) \langle {\bf a}_t \rangle \left[ J \right] \right) \frac{\partial J}{\partial {\bf x}_t}
 + \frac{1}{\beta} \log \sum_{k} \omega_k e^{ - \beta \mathcal{H}_k(J)} \right) dt +  \frac{\partial J}{\partial {\bf x}_t} d {\bf x}_t
\eeq
where  $ \langle {\bf a}_t \rangle_{J} $ is given by Eq.(\ref{a_mean_pi}). The backward SDE (\ref{backward_SDE}) is thus re-written in terms of observables $ {\bf x}_t $ and $ d {\bf x}_t $ (which are given by available data with offline learning), and values of the cost function $ J $ and its derivatives along realized paths of  $ {\bf x}_t $. Note that Eq.(\ref{backward_X}) can be interpreted as the Hamilton-Jacobi (HJ) equation with the inverted time $ t \rightarrow - t $:
\beq
\label{HJ_eq}
d J  =  \mathcal{H}_{HJ} \left[ {\bf x}_t, \xdot_t, J \right] dt
\eeq
where $  \mathcal{H}_{HJ} \left[ {\bf x}_t, \xdot_t, J \right] $ is the `effective' HJ Hamiltonian 
\beq
\label{H_HJ}
 \mathcal{H}_{HJ} \left[ {\bf x}_t, \xdot_t, J \right]  =  
  \frac{1}{\beta} \log \sum_{k} \omega_k e^{ - \beta \mathcal{H}_k(J)}  
  + 
 \left( \xdot_t - \boldmu_0({\bf x}_t) - \boldmu_1({\bf x}_t) \langle {\bf a}_t \rangle \left[ J \right] \right) \frac{\partial J}{\partial {\bf x}_t} + r J 
\eeq
The HJ equation (\ref{HJ_eq}) is the \emph{path-wise} backward recursive relation that enables computing the values of $ J $ and its partial derivatives at time $ t $ from their values at time $ t + dt $ along each path into the future. Unlike the probabilistic soft HJB equation (\ref{Kolmogorov_J_opt_nonlin_GM}) where causality and locality of dynamics are 
\emph{implicit}, in the equivalent path-wise representation of the HJ equation (\ref{HJ_eq}), both locality and causality of dynamics are \emph{explicit}. This may provide better signals for training.   
If the whole solution $ J({\bf x}_t, C_t, t) $ is parameterized by a flexible neural network  $ J_{\theta} ({\bf x}_t, C_t, t) $, the HJ equation (\ref{HJ_eq}) with realized values of $ {\bf x}_t $ and $ d {\bf x}_t $ can be directly used as constraints for learning parameters of the network from available data. A particular way to do it will be presented in the next section.

\section{Learning the soft HJB equation with Deep DOCTR-L}
\label{sect_Neural_PDEs}

\subsection{Loss function for learning  from behavioral data}
\label{sect_Loss_function}

To come up with a loss function that could be used for offline learning, we assume that the solution is encoded into a neural network 
$ J_{\theta}({\bf x}_t, C_t, t)  $, with trainable parameters $ \theta $. The time interval $ [0, T] $ is discretized into a discrete time sequence $ t = t_0, t_1, \ldots t_{T-1}
$ with $ t_0 = 0 $ and $ t_{T-1} = T $ with a time step $ \Delta t $. To ease the notation, we denote the next-step value of $ J({\bf x}_t, C_t,t) $ as $ J({\bf x}_{t+1}, 
C_{t+1}, t+1) $ rather than 
 $ J ({\bf x}_{t+\Delta t}, C_{t + \Delta t}, t + \Delta t) $.  

For a parameterized function $ J_{\theta}({\bf x}_t, C_t, t)  $, a time-discretized version of the HJ equation (\ref{HJ_eq}) for fixed values  $ {\bf x}_t , {\bf x}_{t+1} $ 
is interpreted as a regression:
\beq
\label{backward_X_2_0}
 \Delta J_{\theta}({\bf x}_t, C_t, t) =   \mathcal{H}_{HJ} \left( {\bf x}_t, {\bf x}_{t+1}, J_{\theta} \right) \Delta t +  \nu \varepsilon_t
\eeq
where the Hamiltonian $  \mathcal{H}_{HJ} $ is defined in Eq.(\ref{H_HJ}),  $ \varepsilon_t \sim \mathcal{N}(0, 1) $ and $ \nu^2 $ is the noise variance. 
Note that the noise $ \varepsilon_t $ is introduced here to account for possible inaccuracies of the parametric model $ J_{\theta}({\bf x}_t, t) $, rather than for stochasticity of the dynamics. When the future state is included for learning backward in time as in Eq.(\ref{backward_X_2_0}), stochasticity of the dynamics 
is `frozen' (conditioned on the next-step value $ {\bf x}_{t+1} $).  

Assuming that the function $  J_{\theta}({\bf x}_{t}, C_t,t) $ is known, 
Eq.(\ref{backward_X_2_0}) suggests that the joint probability to observe the  transition $ {\bf x}_t \rightarrow {\bf x}_{t+1} $ along with 
the values  $ J_{\theta}({\bf x}_{t}, C_{t}, t),   J_{\theta}({\bf x}_{t+1}, C_{t+1}, t+1) $ 
at the time step $ [t, t + \Delta t] $ is given by the product of the probability of transition $ {\bf x}_t \rightarrow {\bf x}_{t+1} $ and the probability to observe the 
change $  J_{\theta}({\bf x}_{t+1}, C_{t+1}, t+1) - J_{\theta}({\bf x}_t, C_t, t) $ for given values of  $ {\bf x}_t, {\bf x}_{t+1} $ according to Eq.(\ref{backward_X_2_0}).

Now, learning the parameterized optimal cost functional $ J_{\theta}({\bf x}_t, C_t, t)  $ from \emph{data} implies that transitions  $ {\bf x}_t \rightarrow {\bf x}_{t+1} $ should correspond to the \emph{optimal} policy $ \pi $ given by Eq.(\ref{pi_GM}), which is informative of $ J_{\theta}({\bf x}_t, C_t, t)  $. Importantly, we are \emph{not} given such data, but are rather given samples from the \emph{behavioral} policy $ \pi_0 $. However, we can rely on the likelihood ratio (\ref{likelihood_ratio}) to express probabilities of transitions under the (yet unknown) optimal policy $ \pi $ in terms of probabilities of transitions the behavioral policy $ \pi_0 $ that \emph{are} observed in the data.  To this end, we set  $  \boldmu^{(1)}({\bf x}_t, t) =  \boldmu_0({\bf x}_t) + \boldmu_1({\bf x}_t) \langle {\bf a}_t \rangle \left[ J \right] $ and $  \boldmu^{(0)}({\bf x}_t, t) =  \boldmu_0({\bf x}_t) + \boldmu_1({\bf x}_t) \langle {\bf a}_t \rangle_0  $, where $ \langle {\bf a}_t \rangle_0 $ is the expected action under the behavioral policy (\ref{pi_0_GM}). With these specifications, Eq.(\ref{likelihood_ratio}) reads    
\beq
\label{prob_x1_x2_2}
\frac{ P^{(J)} ( {\bf x}_{t+\Delta t} | {\bf x}_t)}{  P^{(0)} ( {\bf x}_{t+\Delta t} | {\bf x}_t)} 
 = e^{-  \Delta  S({\bf x}_t, {\bf x}_{t+\Delta t}, J_{\theta}) }
 \eeq
where 
\beq
\label{delta_S_2}
\Delta S({\bf x}_t, {\bf x}_{t+\Delta t}, J_{\theta}) =   
\sum_{i} \frac{ \left[ \boldmu_{1}({\bf x}_t) 
\langle {\bf a}_{t} \rangle [J_{\theta}^{-}] \right]_i }{\sigma_i^2}  
\left[ 
\left( \boldmu_{0}({\bf x}_t) + \frac{1}{2} \boldmu_{1}({\bf x}_t)  \langle {\bf a}_{t} \rangle [J_{\theta}^{+}] ) \right)  \Delta t 
- {\bf x}_{t+\Delta t} + 
 {\bf x}_{t} 
\right]_i
\eeq
and $ \langle {\bf a}_{t} \rangle [J_{\theta}^{\pm}] : =  \langle {\bf a}_{t} \rangle [J_{\theta}]  \pm  \langle {\bf a}_{t} \rangle_0 $. 
Note that $ \Delta S({\bf x}_t, {\bf x}_{t+\Delta t}, J_{\theta}) $ in Eq.(\ref{delta_S_2}) is 
a convex function of $ \boldmu_{1}({\bf x}_t)  \langle {\bf a}_{t} \rangle [J_{\theta}] $ with
\bea
\label{min_delta_S}
&& \arg\min_{J_{\theta}}   \Delta S({\bf x}_t, {\bf x}_{t+\Delta t}, J_{\theta}) := \boldmu_1  \langle {\bf a}_{t} \rangle [J_{\theta}]^{(min)} =  \xdot_t - \boldmu_0({\bf x}_t,t) 
\nonumber \\ 
 && \min_{J_{\theta}}   \Delta S({\bf x}_t, {\bf x}_{t+\Delta t}, J_{\theta}) = - \frac{1}{2} \left( \xdot_t- \boldmu_0({\bf x}_t,t) \right)^2
\eea
 and thus the likelihood ratio (\ref{prob_x1_x2_2}) is obviously bounded as a functional of $ J_{\theta} $.
Quantities  $ \langle {\bf a}_{t} \rangle [J_{\theta}^{\pm}]  $ can be computed using Eq.(\ref{a_mean_pi}) as follows:
\beq
\label{a_pm}
 \langle {\bf a}_{t} \rangle [J_{\theta}^{\pm}]  = 
  \sum_{k=1}^{K} \left(\omega_k\left[ J_{\theta} \right]  \frac{{\bf u}_k({\bf x}_t,t) - \beta \sigma_k^2 \boldmu_1({\bf x}_t,t) 
 \frac{\partial J_{\theta}}{\partial {\bf x}_t}}{ 1 + \beta \sigma_k^2 c_1 ({\bf x}_t) \frac{\partial J_{\theta}}{ \partial C_t}} \pm \omega_k {\bf u}_k({\bf x}_t,t) \right)
 \eeq
 The log-likelihood for the unavailable data corresponding to the optimal policy can now be expressed in terms of the likelihood of available \emph{behavioral} data using the analytical likelihood ratio (\ref{prob_x1_x2_2}) as follows:
\beq
\label{LL_PDE}
\mathcal{LL}(\theta) = \log \prod_{t=0}^{T-1}   P^{(0)} ( {\bf x}_{t+\Delta t} | {\bf x}_t) \frac{ P^{(J)} ( {\bf x}_{t+\Delta t} | {\bf x}_t)}{  P^{(0)} ( {\bf x}_{t+\Delta t} | {\bf x}_t)} \exp\left( - \frac{ \left( \Delta J_{\theta}({\bf x}_{t}, t)  -  \mathcal{H}_{HJ} \left( {\bf x}_t,  {\bf x}_{t+\Delta t}, J_{\theta} \right) \Delta t 
\right)^2}{2 \nu^2} \right)
\eeq  
Using Eq.(\ref{prob_x1_x2_2}), flipping the sign and rescaling by $ \nu^2 $, the empirical negative log-likelihood obtained with trajectories $  {\bf x}_{t}^{(n)} $ (with $ n=1, \ldots, N $) corresponding to the behavioral policy $ \pi_0 $ is as follows:
\bea
\label{NLL_PDE}
-   \mathcal{LL}(\theta)  = \frac{1}{N} \sum_{n=1}^{N} \sum_{t=0}^{T-1} &&  \hskip-0.9cm \left[  \frac{1}{2} \left( \Delta J_{\theta}({\bf x}_{t}^{(n)}, C_t^{(n)}, t)  -  \mathcal{H}_{HJ} \left( 
{\bf x}_t^{(n)}, {\bf x}_{t+\Delta t}^{(n)}, J_{\theta} \right) \Delta t \right)^2 \right. \nonumber \\
& + &  \hskip-0.3cm \left.  \nu^2  \Delta S({\bf x}_t^{(n)}, {\bf x}_{t+\Delta t}^{(n)}, J_{\theta}) \right] 
\eea
where the second term is defined in Eq.(\ref{delta_S_2}). 

The loss function (\ref{NLL_PDE}) is the second main contribution of this paper, which provides a recipe for a data-based solution of the soft HJB equation (\ref{Kolmogorov_J_opt_nonlin_GM}).
This loss function is intuitively appealing as it balances the model loss for fixed trajectories, which is given by the first term, with the
cost of a mismatch between the observed and expected dynamics, given by the 
the second term, with $ \nu^2 $ being the only hyperparameter. 
More specifically, the second term $ \sim  \Delta S({\bf x}_t^{(n)}, {\bf x}_{t+\Delta t}^{(n)}, J_{\theta}) $ in Eq.(\ref{NLL_PDE})
encourages consistency of the model $ J_{\theta}({\bf x}_t, t) $ for the optimal cost with dynamics \emph{jointly} implied by observations \emph{and} the model  $ J_{\theta}({\bf x}_t, t) $ according to Eq.(\ref{delta_S_2}). 
The loss function (\ref{NLL_PDE}) has a theoretical low bound which is implied by Eq.(\ref{min_delta_S}).

Note that $  \Delta S({\bf x}_t^{(n)}, {\bf x}_{t+\Delta t}^{(n)}, J_{\theta}) $ depends on the derivatives 
$ \partial J_{\theta} / \partial {\bf x}_t $ and $ \partial J_{\theta} / \partial C_t $, and in a limit when these derivatives can be neglected, 
we have  $ \langle {\bf a}_{t} \rangle [J_{\theta}] \rightarrow  \langle {\bf a}_{t} \rangle_0 $. In this limit, Eq.(\ref{min_delta_S}) would simply enforce matching of the model drift $ \boldmu_0({\bf x}_t) + 
 \boldmu_1({\bf x}_t)  \langle {\bf a}_{t} \rangle_0 $ (which is the drift of the behavioral policy $ \pi_0 $) to observed velocities $\xdot_t $. This implies that `physics' (i.e. passing information contained in function derivatives) is critical for ensuring consistency between the neural network model for the cost function $ J_{\theta} $ and 
 the dynamics jointly implied by observations and the model $ J_{\theta} $.

To summarize, by using its own internal interpretation of behavioral data as \emph{optimal} data `tweaked' by the likelihood  ratio (\ref{prob_x1_x2_2}),
the loss function (\ref{NLL_PDE}) enables \emph{learning the optimal policy directly from the data}. This converts the problem of offline RL into a straightforward supervised learning (inference) problem. Note that while the idea of RL as inference is not new, minimization of the loss function (\ref{NLL_PDE}) solves the problem of optimal control in \emph{one step}, without relying on iterative value iteration or policy iteration methods, as is usually done with discrete-time MaxEnt RL  \cite{Levine_2018}. This is because the  the likelihood ratio (\ref{prob_x1_x2_2}) is itself specified in terms of the optimal value function $ J_{\theta} $ and its derivatives.  
  
\subsection{Deep DOCTR-L solver}
\label{sect_deep_DOCTRL}

The loss function (\ref{NLL_PDE}) can now be used to train a neural network representing a parameterized solution $ J_{\theta} ({\bf x}_t, C_t, t) $.
As the loss  (\ref{NLL_PDE}) involves only the observable data along with the unknown function  $ J_{\theta} ({\bf x}_t, C_t, t) $ and its derivatives, 
we use a single neural network to encode the single unknown  $ J_{\theta} ({\bf x}_t, C_t, t) $, and encode the loss  (\ref{NLL_PDE}) by relying on automatic differentiation available via software such as TensofFlow \cite{Amadi_2015} or PyTorch \cite{PyTorch}. 

In this sense, our algorithm that we call Deep DOCTR-L is analogous to the working of PINNs who similarly encode the solution into a single neural network. Also similarly to PINNs, our method gives a solution at \emph{arbitrary} arguments. \emph{Unlike} a PINN that encodes the original PDE, the present approach encodes the loss function (\ref{NLL_PDE})
which is based on \emph{path-wise} (and step-wise) information. This loss function is expected to provide stronger signals for training than encoding of the original PDE, and is designed to work in high dimensions. Paths used for training of Deep DOCTR-L are related to paths that would be observed under the optimal policy by the likelihood ratio {(\ref{prob_x1_x2_2}), and are thus informative of the solution. This is different from PINNs that are typically analyzed on an arbitrary mesh that has no link with the underlying dynamics of the system described by the PDE.  

Another difference from PINNs is that Deep DOCTR-L has only one hyperparameter $ \nu^2 $, and does not engage two different datasets (and an extra hyperparameter) for an interior and a boundary as is done for PINNs. This is because the soft HJB equation, as well as the classical HJB equation, only has a terminal condition, while having natural boundary conditions that need not be explicitly enforced.
The terminal condition on $ J_{\theta}({\bf x}_T, C_T, T) =  U(C_T)  $ (see Eq.(\ref{J_T})) is used for the last-step value $ \Delta J_{\theta}({\bf x}_{T}, C_T, T) $ in the 
loss (\ref{NLL_PDE}).

While sharing similarities with PINNs due to its reliance on a single neural network and operating only with observed data, Deep DOCTR-L also shares some similarities with deep FBSDE methods \cite{Han_2017, Raissi_2018, Zhang_2020} such as the Deep BSDE solver \cite{ Han_2018}. Similarly to the latter, it uses information that goes beyond the soft HJB equation (\ref{Kolmogorov_J_opt_nonlin_GM}) itself, and incorporates path-wise dynamics. 
With our offline RL method, the FBSDE dynamics
are restated as the equivalent path-wise HJ equation (\ref{HJ_eq}) which is formulated directly in terms of observable variables $ {\bf x}_t, {\bf x}_{t+ \Delta t} $.
The HJ dynamics are then enforced as a constraint on a parametric approximation $ J_{\theta} $ in the loss function (\ref{NLL_PDE}). 

Similarly to deep FBSDE methods,  the Deep DOCTR-L solver directly computes the optimal value function $ J_{\theta} $  and the optimal policy (\ref{pi_GM}) in one step. However, differently from the former, it does not 
use additional neural sub-networks to represent the value function gradients. This leads to a better data efficiency.
Another difference is that while deep FBSDE methods are model-based approaches that
rely on simulations of auxiliary stochastic processes according to a known model, Deep DOCTR-L does not use simulations and learns directly from offline data.

As Deep DOCTR-L directly computes the optimal value function $ J_{\theta} $ from data using the supervised learning loss (\ref{NLL_PDE}) without using additional simulated quantities as in deep FBSDE methods, it means that  to become operational, it only needs to implement the `custom' loss function (\ref{NLL_PDE}) 
with a generic neural network-based algorithm for supervised learning. (In addition, it uses a Gaussian mixture model for the behavioral policy $ \pi_0 $, which can be estimated using off-the-shelf scientific software libraries.)
Our implementation uses PyTorch to implement a simple feedforward neural network representing a parameterized optimal value function $ J_{\theta} ({\bf x}_t, C_t, t) $, and relies on automatic differentiation to compute derivatives of the value function that enter the loss  (\ref{NLL_PDE}).

\section{Experiments}
\label{sect_Experiments}

\subsection{Learning in 10 dimensions}
\label{sect_learning_in_10D}

In the first experiment, we simulate behavioral data in a 10-dimensional space $ {\bf X} $ ($ D_x = 10$), with a 5-dimensional action space ($D_a = 5$). Trajectories have 40 steps that uniformly partition the time interval $ t \in [0,T] $ with $ T = 1 $.  The number of simulated trajectories is 10,000. The following specifications of drift and cost functions are used (here $ I_{D_1, D_2} $ stands for a unit matrix of size $ D_1, D_2 $, and $ I_{D} $ stands for a unit vector of size $ D $):
\bea
\label{specs_10D}
&& \boldmu_0({\bf x}_t) =   \boldmu_0^{(0)} +   {\bf x}_t \cdot \boldmu_0^{(1)},  \; \;  \; \boldmu_0^{(0)} = 0.1 I_{D_x}, \; \; 
\boldmu_0^{(1)} = 0.2 I_{D_x,D_x} \nonumber \\
&& \boldmu_1({\bf x}_t) =   \boldmu_1^{(0)} +   {\bf x}_t \cdot \boldmu_1^{(1)}, \; \;  \; \boldmu_1^{(0)} = 0.1 I_{D_x,D_a}, \; \; 
\boldmu_1^{(1)} = 0.2 I_{D_x,D_a} \nonumber \\
&& c_0({\bf x}_t) = c_0  \left| \left|  {\bf x}_t \right| \right|^2, \; \; \;  c_0 = 1.0; \; \; \; 
 c_1({\bf x}_t) = c_1  \left| \left|  {\bf x}_t \right| \right|^2, \; \; \;  c_1 = 5.0 
\eea 
For a behavioral policy $ \pi_0 $, we use a Gaussian mixture with $ K = 2 $ components, where the means $ {\bf u}_k $ are chosen to be independent of $ {\bf x}_t $, and sampled from a uniform distribution on the interval $ [-0.5, 0.5] $ for each one of the $ D_a = 5 $ components of action $ {\bf a}_t $. Variances of the Gaussian components of the policy $ \pi_0 $ are sampled from a uniform distribution on $[0.2, 0.4] $. The weights $ \omega_k $ of the prior $ \pi_0 $ are uniform on $ K $. We use the following values of parameters: $ \beta = 1 $, $ \nu^2 = 100 $, and $ r = 0.03 $.

Giving the form of the prior policy $ \pi_0 $ as just specified, simulation of the `effective' 10-dimensional diffusion is done using Eqs.(\ref{CT_Langevin_controlled_pi_2}), where we use policy $ \pi_0 $ to compute the policy-dependent drift according to Eq.(\ref{drift_sigma_pi}) at each time step.
Fig.~\ref{fig_Simulated_trajs} shows 100 randomly selected trajectories for a randomly selected component of the 10-dimensional diffusion in the $ {\bf X} $ space induced by the behavioral policy $ \pi_0 $ according to such procedure. Note for what follows a limited range of values of $ {\bf x}_t $, which, for the chosen component, vary between approximately 0.02 and 0.45. Once a set of trajectories $ {\bf x}_t $ is simulated in this way, it is updated to a new dataset for extended trajectories $ {\bf y}_t = ({\bf x}_t, C_t) $, where cumulative costs $ C_t $ are obtained using specifications in Eqs.(\ref{specs_10D}). In our case, the extended space $ {\bf Y} $ has $ N = 11 $ dimensions.

\begin{figure}[ht]
\begin{center}
\includegraphics[
width=80mm,
height=55mm]{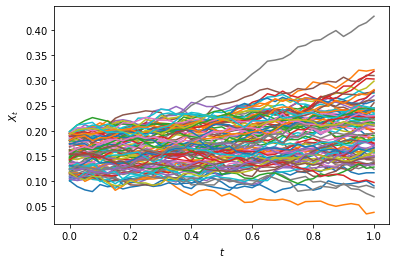}
\caption{Simulated trajectories: 100 randomly selected time series for a randomly selected component of a 10-dimensional process $ {\bf x}_t $ obtained with Eq.(\ref{CT_Langevin_controlled_pi_2}) using the behavioral policy $ \pi_0 $. Note that all simulated values lie in the band [0.02, 0.45], with no available data beyond this range. } 
\label{fig_Simulated_trajs}
\end{center}
\end{figure}

The optimal value function $ J_{\theta}({\bf y}_t, t) $ is approximated by a feedforward neural network with 3 hidden layers, each having 100 neurons with a softplus activation function, implemented using PyTorch 1.8.0. Training is done using the Adam optimization with mini-batches of size 256, with a L2 regularization for weights of the neural network with parameter 0.001. Training for 10 dimensions takes about 8 min on 
a Mac laptop with a 2.5 GHz Intel Core i7 processor.
The results of training along with a schedule for the learning rate are shown in Fig.~\ref{fig_training_10D}.

\begin{figure}[ht]
\begin{center}
\includegraphics[
width=130mm,
height=55mm]{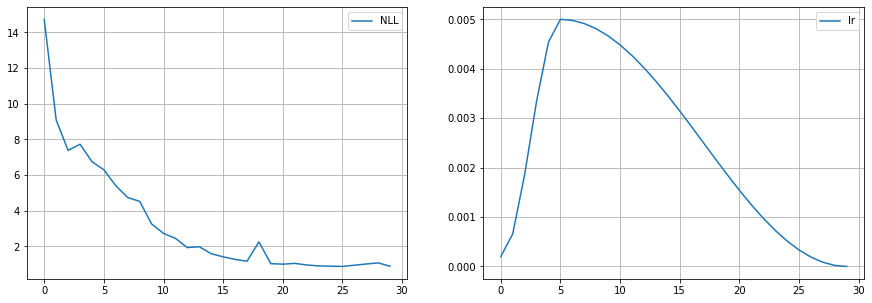}
\caption{Training performance of Deep DOCTR-L on 10-dimensional data. On the left: the negative log-likelihood as a function of the training epoch.
On the right: the learning rate schedule as a function of the epoch. 
} 
\label{fig_training_10D}
\end{center}
\end{figure}

Once the training is completed, the learned value function  $ J_{\theta}({\bf y}_t, t) $, along with its derivatives with respect to inputs
$ \left( t, {\bf x}_t, C_t \right) $, are available for arbitrary inputs. The optimal policy $ \pi $ is then computed using Eqs.(\ref{pi_GM}) and (\ref{means_vars}). To assess the quality and behavior of the learned optimal policy $ \pi $, Fig.~\ref{fig_GM_means_10D} shows the state dependence for all $ D_a = 5 $ dimensions of the Gaussian means $ {\bf u}_k \left[ J \right] $ for both components of the Gaussian mixture policy $ \pi $, as functions of the same randomly chosen component of a 10-dimensional vector $ {\bf x}_t $ that was used above in Fig.~\ref{fig_Simulated_trajs}.
 
\begin{figure}[ht]
\begin{center}
\includegraphics[
width=130mm,
height=55mm]{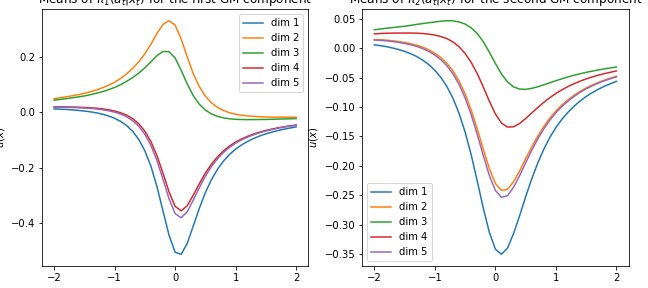}
\caption{The state-dependence of means $ {\bf u}_k $ of the GM optimal policy $ \pi $ for all dimensions of the action space, as a function of a randomly chosen component of a 10-dimensional state variable $ {\bf x}_t $.  On the left: The first GM component. On the right: the second GM component. All means asymptotically approach a constant value with a vanishing $ x $-gradient.
} 
\label{fig_GM_means_10D}
\end{center}
\end{figure}

The solution for the optimal policy shown in Fig.~\ref{fig_GM_means_10D} warrants some comments. First, as the prior means $ {\bf u}_k $ were chosen to be state-independent, this means that the state dependence of the optimal means $ {\bf u}_k \left[ J 
\right] $ is solely due to the state dependence of $ J_{\theta} $ and its gradients.
All profiles for $ {\bf u}_k(x) $ are very smooth, meaning that the optimal solution gives rise to both a smooth value function \emph{and} smooth gradients.\footnote{Note that such a smooth behavior was observed when using 
the softplus activation function for hidden layers. Experiments performed using alternative specifications including the $ \tanh $ and $ \sin $ activations produced more noisy profiles for the same data, with discontinuities of derivatives, i.e. a non-smooth behavior of functions  $ {\bf u}_k(x) $ at certain values of $ x $.}
Second, all dimensions of the optimal policy have a mode at approximately the same value of the argument. This is presumably related to the symmetry of the cost $ c_1 ({\bf x}_t, {\bf a}_t ) =  c_0({\bf x}_t) + c_1({\bf x}_t)  || {\bf a}_t ||^2 $ under rotations in the $ {\bf a}_t $ space.  

The third, and most important observation is that all means $ {\bf u}_k(x) $ approach constant (and similar) values at large negative or large positive values of $ x $. Recall from 
Fig.~\ref{fig_Simulated_trajs} that only a limited range of values of $ x $ between 0.02 and 0.45 was originally provided in the training dataset.
This suggests that the trained optimal value function  $ J_{\theta}({\bf y}_t, t) $ is able not only to \emph{interpolate} between input values encountered in a training dataset, but also to \emph{extrapolate} beyond demonstrated ranges of inputs. Clearly, the ability to extrapolate with asymptotically flat solutions critically depends on the ability of the network to learn not only the value function but also its gradients.  

\subsection{Learning in 100 dimensions}
\label{sect_learning_in_100D}

The experimental setting for the offline learning of control in a 100-dimensional state space is similar to the setting of the previous experiment. The action space dimension is 
kept at $ D_a = 5 $, and the number of simulated trajectories is 10,000, the same as in the 10-dimensional case above. All parameters are kept the same except $ 
\boldmu_0^{(1)}, \, \boldmu_1^{(1)}, \nu^2 $, which are all divided by 10 relative to their values in the previous 10-dimensional case to keep the relative order of different terms in the loss function to be approximately the same. In addition, a higher value of $ \beta = 5 $ is used in this example.\footnote{Lower values $ \beta = 1 $ or $ \beta = 3 $ provide visually similar results.}

The results of training along with a schedule for the learning rate are shown in Fig.~\ref{fig_training_100D}. The CPU compute time for this example was 
approximately 45 min for training with 30 epochs.

\begin{figure}[ht]
\begin{center}
\includegraphics[
width=130mm,
height=55mm]{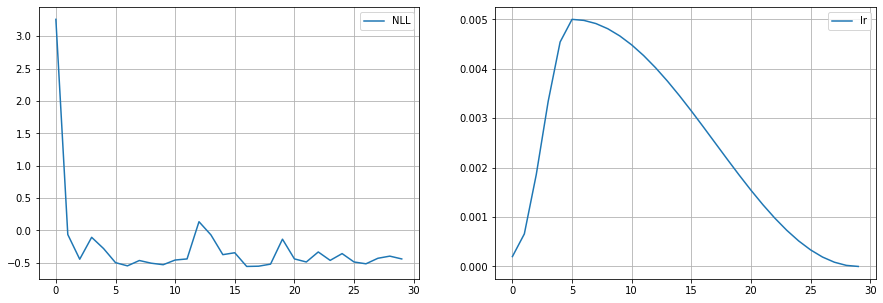}
\caption{Training performance of Deep DOCTR-L on 100-dimensional data. On the left: the negative log-likelihood as a function of the training epoch.
On the right: the learning rate schedule as a function of the epoch. 
} 
\label{fig_training_100D}
\end{center}
\end{figure}

Fig.~\ref{fig_GM_means_100D} shows the state dependence for all $ D_a = 5 $ dimensions of the Gaussian means $ {\bf u}_k \left[ J \right] $ for both components of the Gaussian mixture policy $ \pi $, for a randomly chosen component of a 100-dimensional vector $ {\bf x}_t $. The resulting behavior is very similar to the one found for the 10-dimensional case.
Again, all solutions are very smooth, and asymptotically approach constant values. Due to a higher dimensionality of data, asymptotic levels are reached for larger absolute values of the input $ x $, relatively to the previous 10-dimensional case. Again, this 100-dimensional example shows that Deep DOCTR-L is able to learn and \emph{extrapolate} optimal policies using a moderate amount of training data.\footnote{10,000 training trajectories that are used here to learn the optimal policy in a 100-dimensional case is of the same order of magnitude as numbers of training samples typically used with PINN networks for solving PDEs in low dimensions.}. 
 
\begin{figure}[ht]
\begin{center}
\includegraphics[
width=130mm,
height=55mm]{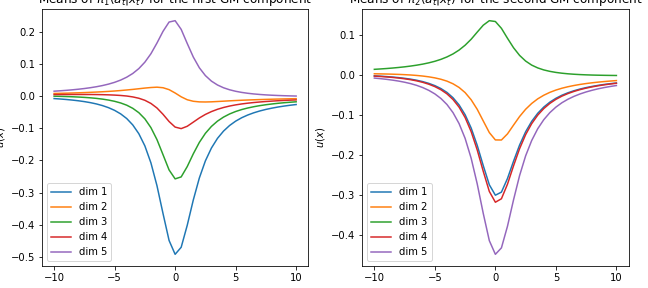}
\caption{The state-dependence of means $ {\bf u}_k $ of the GM optimal policy $ \pi $ for all dimensions of the action space, as a function of a randomly chosen component of a 100-dimensional state variable $ {\bf x}_t $.  On the left: The first GM component. On the right: the second GM component. All means asymptotically approach a constant value with a vanishing $ x $-gradient.
} 
\label{fig_GM_means_100D}
\end{center}
\end{figure}

\section{Summary}
\label{sect_Summary}

Numerical experiments conducted above for high-dimensional control problems using examples with state spaces of dimensions 10 and 100 
demonstrate the ability of the Deep DOCTR-L solver to learn from offline data using a moderate amount of data. 

Unlike other algorithms for offline RL, Deep DOCTR-L completely avoids computationally intensive value iteration or policy iteration methods.
This is due to
the analytical expression (\ref{prob_x1_x2_2}) for the likelihood ratio between transitions observed using the behavioral policy $ \pi_0 $ 
and transitions that would be observed using the optimal policy $ \pi $. This expression enables writing the loss function
(\ref{NLL_PDE}) directly in terms of the optimal value function  $ J_{\theta}({\bf y}_t, t) $ and its derivatives. 
This demonstrates the benefits of the suggested SciPhy RL approach that reduces distributional offline continuous-time RL to solving PDEs from data.
The SciPhy RL approach thus achieves the goal of reformulation of offline RL as 
a supervised learning task with the loss (\ref{NLL_PDE}). The algorithm employs a single neural network which is able to extrapolate the optimal value function and optimal policy to input values that may be very different from those encountered in the training data. 

While the focus of this work was on the theoretical formulation of the Deep DOCTR-L method, initial experiments conducted here suggest that compute times 
with this methods are quite manageable (measured in minutes) even with a single CPU in high-dimensional state spaces up to a 100-dimensional case. 
Of course, further experimental tests would be needed to explore the performance for different environments and cost models, which is left here for a future work.


%
%

\end{document}